%% file: root.tex
\documentclass{ifacconf}

\usepackage[utf8]{inputenc}
\usepackage{graphicx}      
\usepackage{amssymb}
\usepackage{float}
\usepackage{subfigure}
\usepackage{array}
\usepackage{booktabs}
\usepackage{calc}
\usepackage{multirow}
\usepackage{natbib}        
\usepackage[firstpage=true, placement=top]{background}
\backgroundsetup{contents={\parbox{\textwidth+1cm}{\copyright 2021 the authors. This work has been accepted to IFAC for publication under a Creative Commons Licence CC-BY-NC-ND.}}, color=black, opacity=1, scale=0.9, vshift=-10pt}
\usepackage{tikz}
\usetikzlibrary{shapes,arrows,plotmarks,positioning, decorations.pathreplacing}
\begin{document}
\begin{frontmatter}

\title{Learning to Model the Grasp Space of an Underactuated Robot Gripper Using Variational Autoencoder} 

\author[First]{Clément Rolinat, Mathieu Grossard,} 
\author[Second]{Saifeddine Aloui, Christelle Godin} 

\address[First]{Université Paris-Saclay, CEA, List, F-91120, Palaiseau, France (e-mail: \{clement.rolinat, mathieu.grossard\}@cea.fr).}
\address[Second]{Université Grenoble Alpes, CEA, Leti, F-38000, Grenoble, France (e-mail: \{saifeddine.aloui, christelle.godin\}@cea.fr)}

\begin{abstract}                
Grasp planning and most specifically the grasp space exploration is still an open issue in robotics. This article presents a data-driven oriented methodology to model the grasp space of a multi-fingered adaptive gripper for known objects. This method relies on a limited dataset of manually specified expert grasps, and uses variational autoencoder to learn grasp intrinsic features in a compact way from a computational point of view. The learnt model can then be used to generate new non-learnt gripper configurations to explore the grasp space.
\end{abstract}

\begin{keyword}
multi-fingered gripper, grasp space exploration, variational autoencoder
\end{keyword}

\end{frontmatter}

\section{Introduction}
\input{introduction_short.tex}

\section{Problem Statement \& Framework}
\label{sec:statement}
\input{prob_statement.tex}

\section{Method Description}
\label{sec:method_desc}
\input{method_desc.tex}

\section{HGG Tuning to Model Efficiently the Grasp Space}
\label{sec:vae}
\input{vae_tuning.tex}

\section{Conclusion}
\input{conclusion.tex}

\bibliography{references.bib}             

\end{document}

%% file: introduction_short.tex
Grasping is fundamental in most of the industrial manufacturing processes such as pick-and-place, assembly or bin picking tasks. The grasp planning question is still an active research topic. It aims at finding a gripper configuration that allows to grasp an object reliably. From a geometrical point of view, the chosen grasp configuration needs to be kinematically reachable and collision-free with respect to the environment, while, from a dynamics point of view, the grasp needs to ensure object stability and resistance against external perturbations. Finding such a grasp configuration requires to explore the grasp space, that is the subset of gripper configurations that effectively grasp the object. Thus, grasp planning is both object dependent and robot hardware dependent. Taking into account those constraints during the exploration is not straightforward, as objects can have sophisticated shapes, and gripper-arm combination can have complex kinematics. 

This is even more true for underactuated or compliant multi-fingered gripper, for which the adaptive under-actuated system generates object-dependent grasp configurations. This type of grippers are often chosen for grasping tasks \citep{townsend_barretthand_2000}. Indeed, such architecture allows to reduce the controller complexity by reducing the number of controlled degrees of freedom, while retaining sufficient kinematic abilities. Moreover, it tends toward producing robust grasps by their mechanical structure.

The grasp planner should be able to find in the high dimensional and highly constrained grasp space a configuration that fulfills a given criterion. There are two main ways to achieve this: analytic approaches and data-driven approaches \citep{sahbani_overview_2012}. Analytic approaches rely on an analytic description of the grasping problem (see \cite{berenson_grasp_2007}, \cite{roa_grasp_2008}, \cite{xue_grasp_2007}). Data-driven approaches depend on machine learning methods to predict grasps from object depth map or point cloud (see \cite{zhao_grasp_2020}, \cite{pinto_supersizing_2015}, \cite{depierre_jacquard:_2018}, \cite{levine_learning_2018}, \cite{mahler_dex-net_2017}).

A shared issue is the grasp dataset creation, that is the grasp space exploration. A variety of high quality grasps needs to be discovered by exploring the space of possible grasp configurations. There are two main approaches regarding this exploration \citep{xue_grasp_2007}: contact point approaches, and gripper configuration approaches. In the first case the grasp space exploration comes down to test various combinations of contact point locations on the object surface. However there is no guarantee that a given combination is a priori kinematically admissible for a given gripper, and the inverse kinematics can even be intractable for underactuated or adaptive grippers. In the second case, the grasp space is explored by testing several gripper spatial configurations. This is more suited for underactuated grippers. Nevertheless, there is no assurance that a given gripper configuration is a priori in contact with the object without realizing extensive simulation trials beforehand.

To circumvent the dimensionality issue related to the huge size of the grasp space, numerous contact point approaches limit their search to fingertip contacts (see \cite{roa_grasp_2008}, \cite{zhao_grasp_2020}), and gripper configuration approaches often use a bi-digital gripper and limit their search to planar grasps (see \cite{pinto_supersizing_2015}, \cite{depierre_jacquard:_2018}, \cite{levine_learning_2018}, \cite{mahler_dex-net_2017}). For more complex grippers, a human input is often required. For example, in \cite{santina_learning_2019}, the authors identified a set of ten grasp primitives from human examples, and reduced the grasp space to those primitives only. In \cite{choi_learning_2018}, the authors proposed to limit the search space by discretizing it.

This article presents a method to model the grasp space of a multi-fingered and underactuated gripper for known objects using a variational autoencoder. It relies on a limited set of human-chosen primitive gripper configurations. This model can then be used to generate new gripper configurations that are likely to belong to the grasp space. This allows to explore the grasp space considering inspiration from already specified human-based complex grasps strategies.

The section \ref{sec:statement} is dedicated to the problem statement and a presentation of the framework useful for our work. Then, in section \ref{sec:method_desc} the method used to model the grasp space is described. Finally, in section \ref{sec:vae} we show how to tune the variational autoencoder hyperparameters to learn the best grasp space model. To conclude, this work is discussed and the planned future works are presented.

%% file: prob_statement.tex
\subsection{Simulation Setup}

The three-fingered gripper considered in the following has an underactuated and adaptive behavior that allows its natural adaptation to the object geometry, without the need to carefully control each joint, thus increasing the robustness of the grasp.

This gripper has two joints on each finger and one actuator per finger to control both joints. The second (distal) phalanx starts moving when the applied effort on the finger is above a given force threshold. A fourth actuator allows to control the spread angle $\theta$ between two fingers (see Fig. \ref{fig:gripper_info}).

This gripper is mounted as end effector of a six degrees of freedom industrial robot arm.

The simulation setup described above is implemented with Gazebo simulator \citep{koenig_design_2004}. A picture of this simulated setup is displayed in Fig. \ref{fig:gripper_info}.

\subsection{Problem Statement}
\label{subsec:prob_statement}

An object is placed on a table in the workspace of the considered robotic setup. It is assumed that the object geometry is known, as well as the pose in the scene of its associated frame $F_{obj}$ thanks to exteroceptive vision system such as in \cite{drost_model_2010} for example.

\begin{figure}[ht!]
\centering
\subfigure[Pose of the frame $F_{grip}$ used to locate the end-effector relatively to $F_{obj}$. \label{subfig:gripper_frame}]{\includegraphics[width=0.632\linewidth, keepaspectratio]{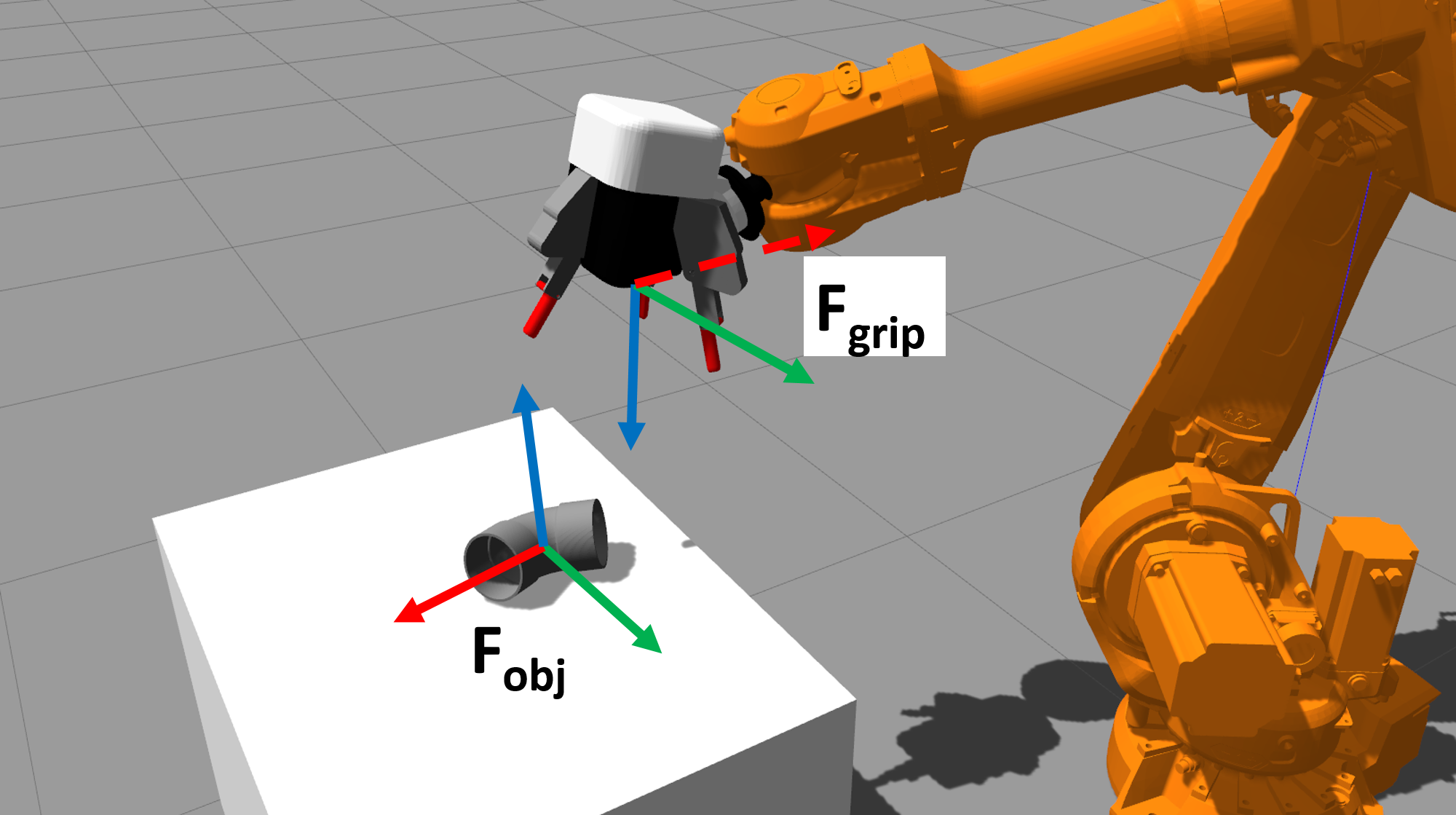}}
\subfigure[Spread angle $\theta$. \label{subfig:gripper_spread}]{\includegraphics[width=0.344\linewidth, keepaspectratio]{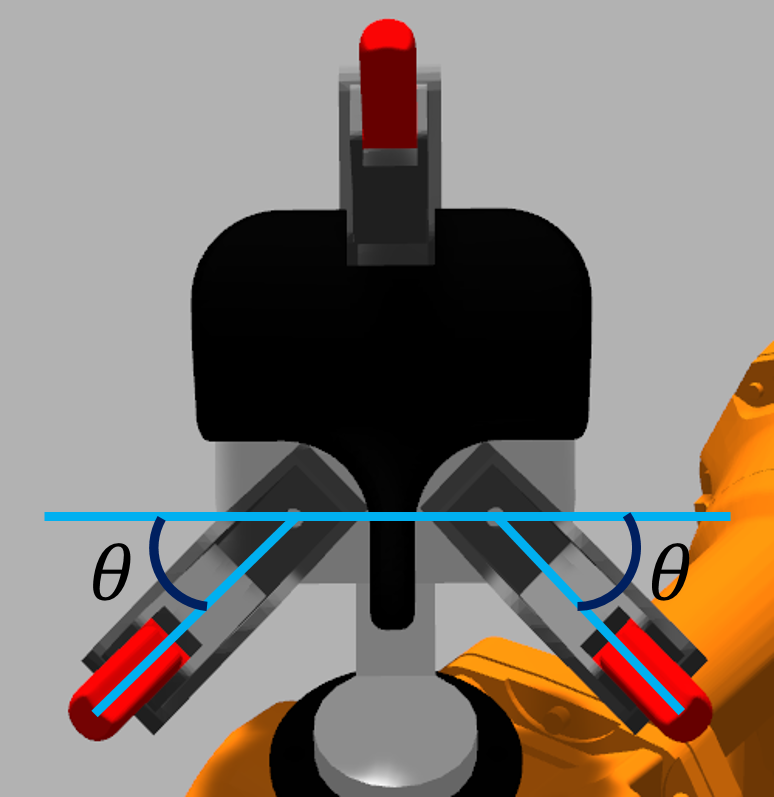}}
\caption{Gripper frame and spread angle.}
\label{fig:gripper_info}
\end{figure}

In the following, a grasp configuration is a gripper configuration that is able to grasp the object without colliding with the table. It is defined as follows by eight parameters:
\begin{itemize}
\item the pose of the gripper frame $F_{grip}$,
\[
(x, y, z, q_x, q_y, q_z, q_w) \in \mathbb{R}^{3} \times SO(3)
\]
with the orientation expressed in quaternion convention;
\item the spread angle $\theta$, as shown in Fig. \ref{fig:gripper_info}. 
\end{itemize}

The dimensionality of this configuration space is high (seven dimensions), but this allows to fully leverage the grasping ability and kinematic potential of the gripper. Thus, the grasp space is a subset of this gripper configuration space, with an additional constraint that every gripper configuration is able to grasp the object without colliding with the table. The goal is to model and explore this grasp space.

To locate the gripper, a dedicated frame $F_{grip}$ situated between the fingers in front of the palm is used. This frame is displayed in Fig. \ref{fig:gripper_info}. Gripper poses are expressed relatively to the object frame $F_{obj}$, in order to be invariant to object poses.

\subsection{Variational Autoencoders}

Variational autoencoders (VAE) are a derivative of classic autoencoders. In addition to learning a compressed representation of the training data as a classic autoencoder, a VAE allows to generate consistent data from its latent space reliably.

The goal of a VAE is to infer the latent variables that are behind the training dataset, so that a latent space with fewer dimensions than the original space can be created. Sampling in it will generate a new data distribution that resembles to the training dataset one. In this work, this allows to create a model of the grasp space, from which new gripper configurations can be generated.

In a VAE, among other features, a supplementary term is added in the loss function: the Kullback-Leibler (KL) divergence \citep{kingma_auto_2014}. This term helps the data to be represented as a normal distribution in its latent space, and thus regularizing it.

%% file: method_desc.tex
\begin{figure*}
\centering
\input{scheme_architecture}
\caption{HGG architecture. In blue the input layers, in green the hidden neural networks (NN) and in red the output layers. The hidden NN inner layers are fully connected layers, with hyperbolic tangent activation function. The latent space dimension, that is the number of latent variables, is $n$. The main encoding and main decoding NN have symmetrical inner architecture. The supplementary input for the tabletop equation ensures that the generated grasp depends on it \citep{sohn_learning_2015}. This architecture is implemented with Tensorflow \citep{tensorflow2015-whitepaper} and Keras \citep{chollet2015keras} python libraries.}
\label{fig:archi}
\end{figure*}
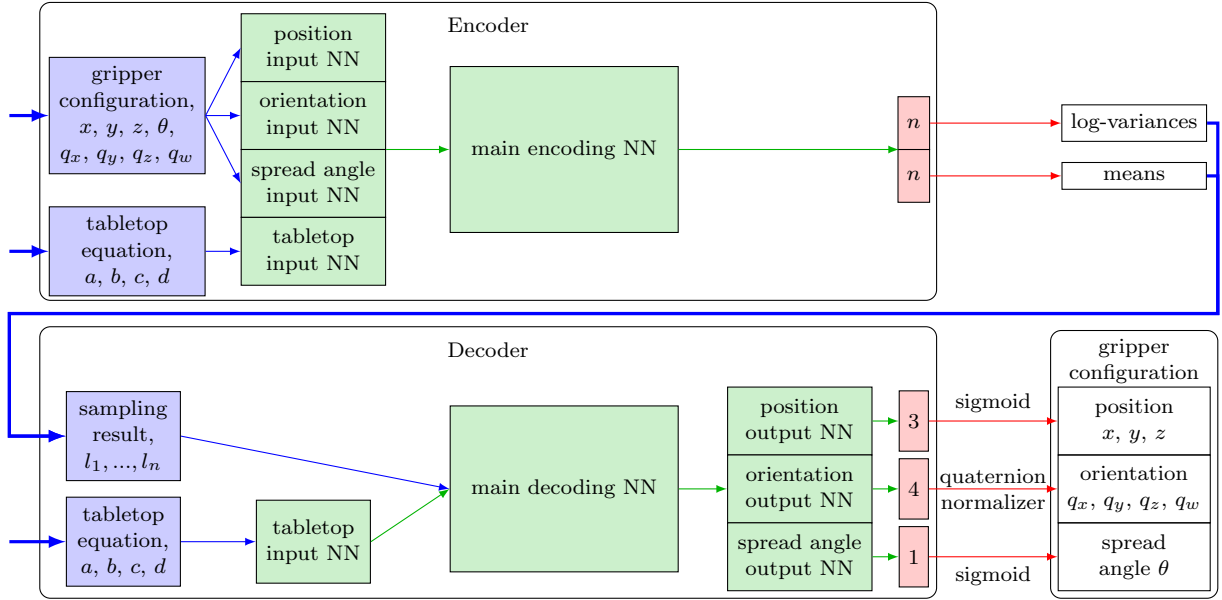

The proposed method has two main steps : an object depend primitive dataset building step, and a Human-initiated Grasp Generator VAE (HGG) training step.

\subsection{Primitive Grasp Dataset}

To leverage the human ability to find gripper configurations belonging to the grasp space, an object dependent primitive grasp dataset is built. A primitive grasp is a handcrafted gripper configuration, with its pose and spread angle human-chosen so that it is collision free and likely to grasp the object. The spread angle is chosen between four discrete values corresponding to main gripper internal layouts: $\theta = 0$, $\theta = \pi/6$, $\theta = \pi/4$, and $\theta = \pi/2$.

In this work, such primitives are gathered on three different objects:
\begin{itemize}
 \item a connector bent pipe
 \item a pulley
 \item a small cinder block
\end{itemize}

\begin{figure}[htb]
\setlength{\tabcolsep}{0pt}
\renewcommand{\arraystretch}{0}
\centering
\begin{tabular}{>{\centering\arraybackslash}m{1.8cm} c c c}
bent pipe & \raisebox{-0.5\height}{\includegraphics[width=0.25\linewidth, keepaspectratio]{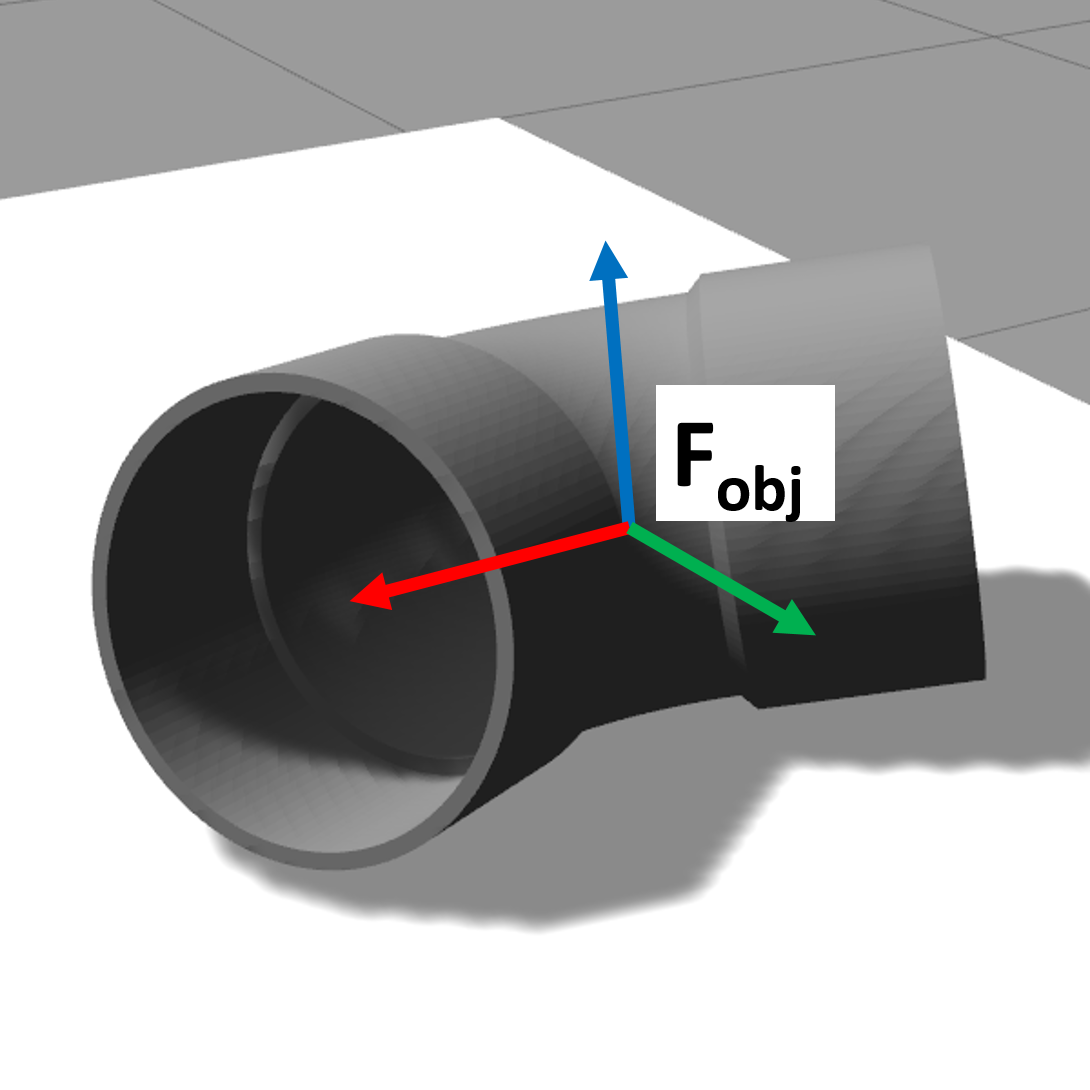}} &
\raisebox{-0.5\height}{\includegraphics[width=0.25\linewidth, keepaspectratio]{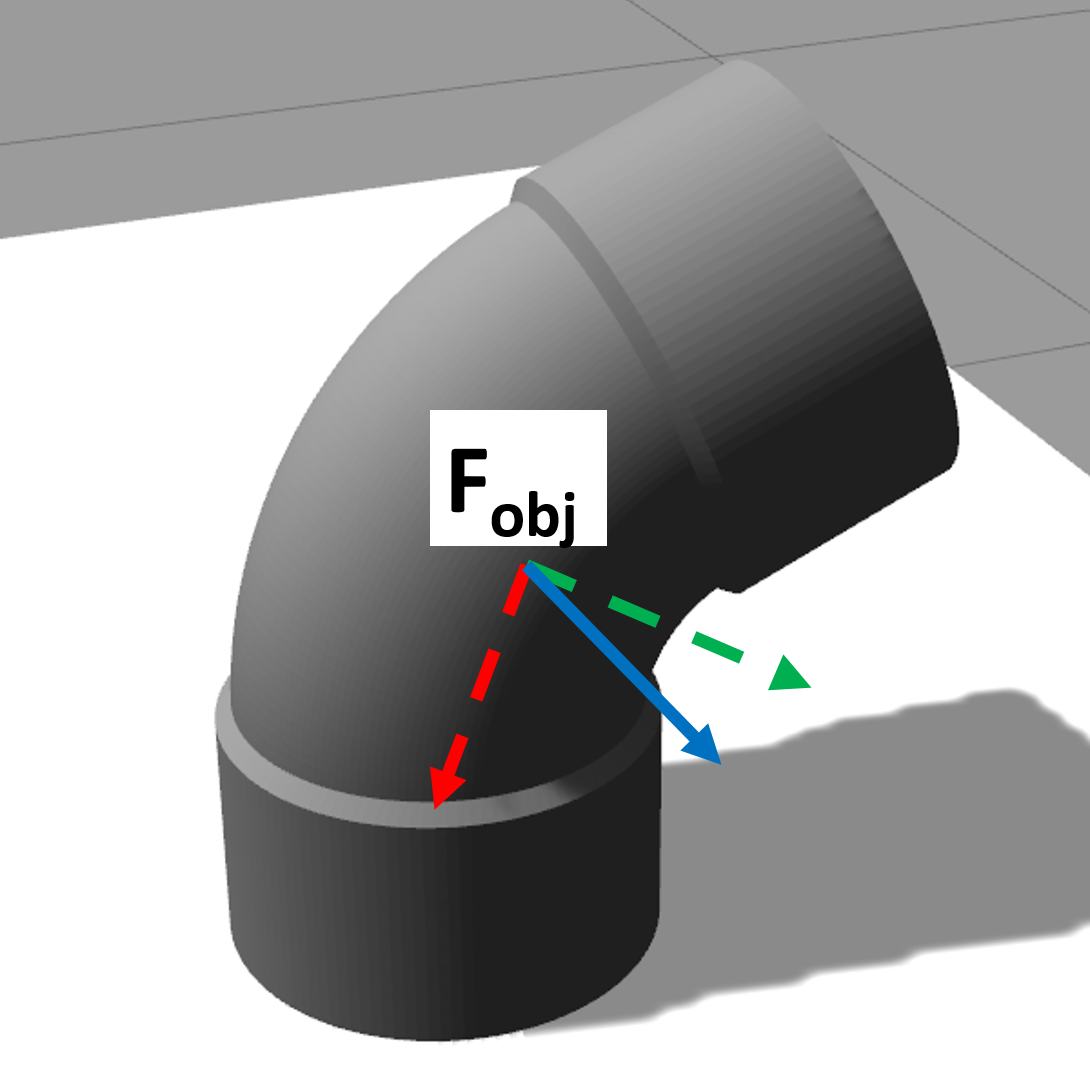}} &
\\
cinder block & \raisebox{-0.5\height}{\includegraphics[width=0.25\linewidth, keepaspectratio]{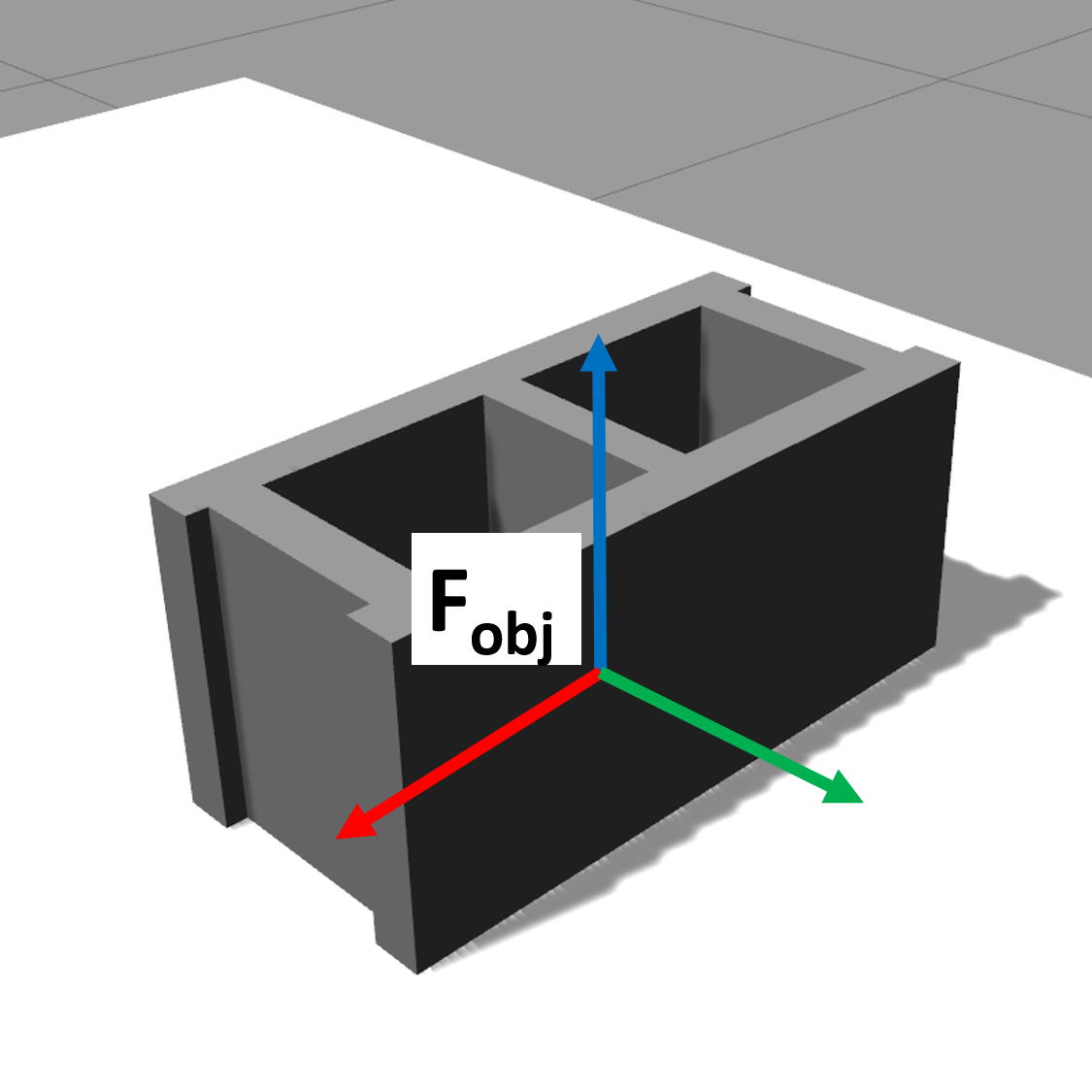}} &
\raisebox{-0.5\height}{\includegraphics[width=0.25\linewidth, keepaspectratio]{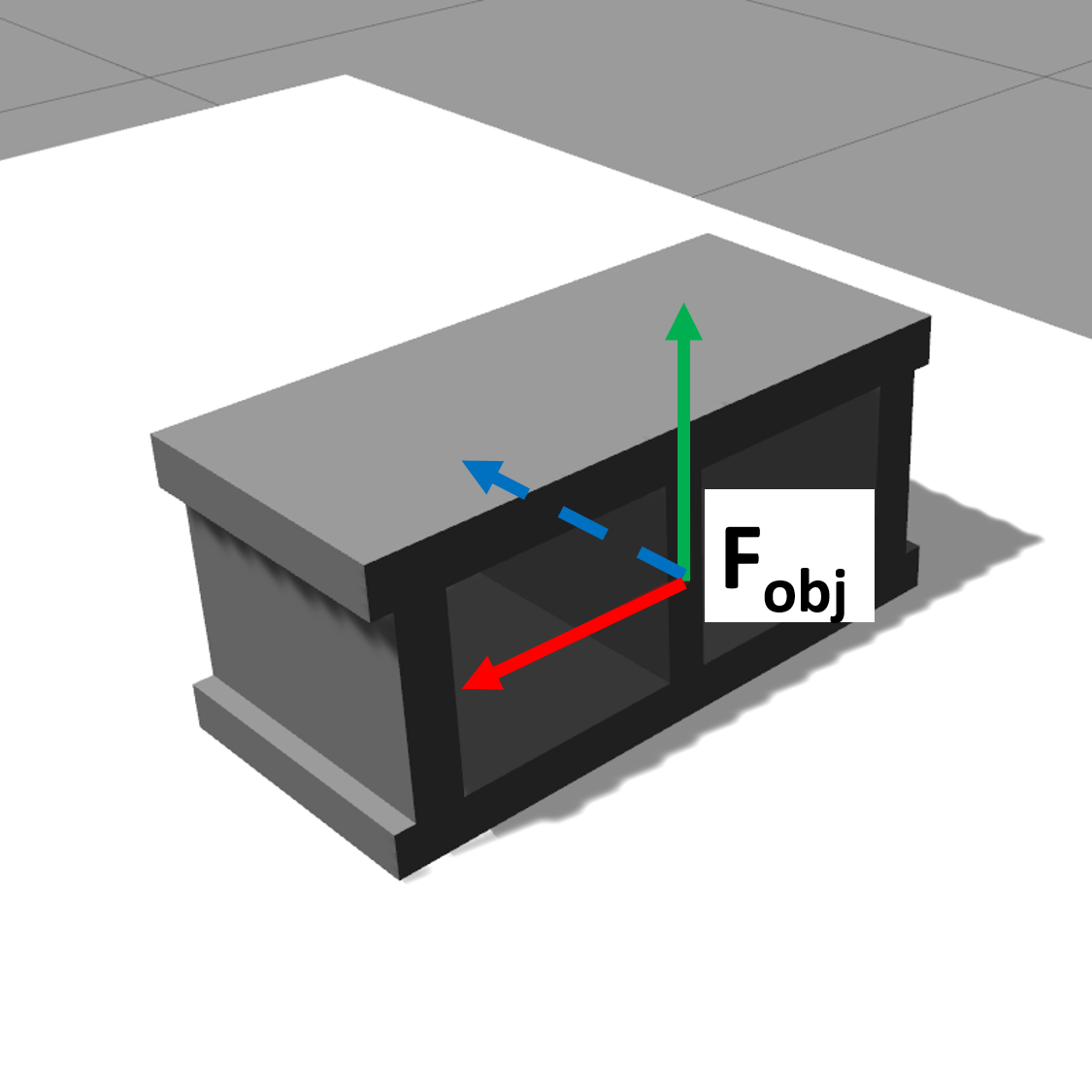}} &
\raisebox{-0.5\height}{\includegraphics[width=0.25\linewidth, keepaspectratio]{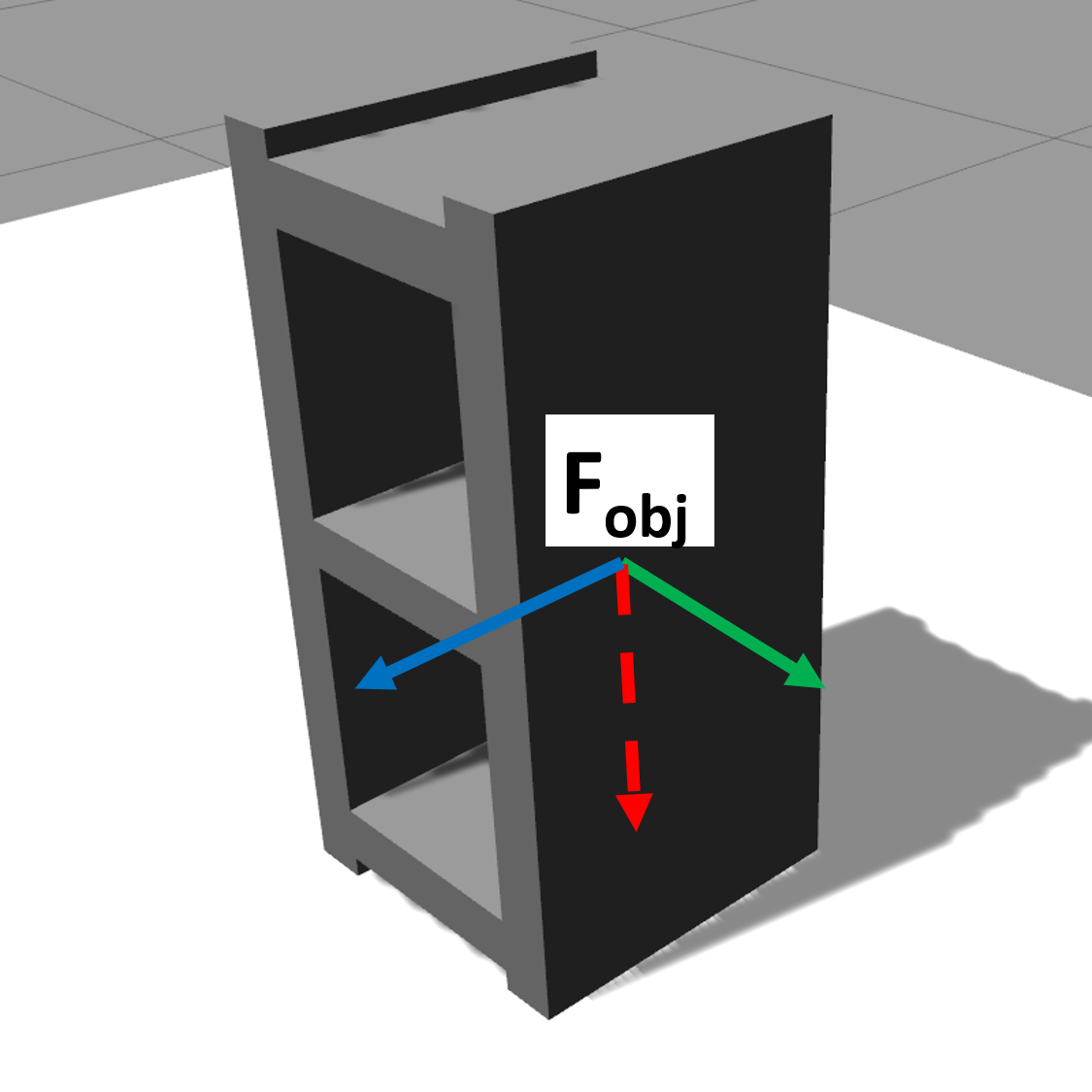}}
\\
pulley & \raisebox{-0.5\height}{\includegraphics[width=0.25\linewidth, keepaspectratio]{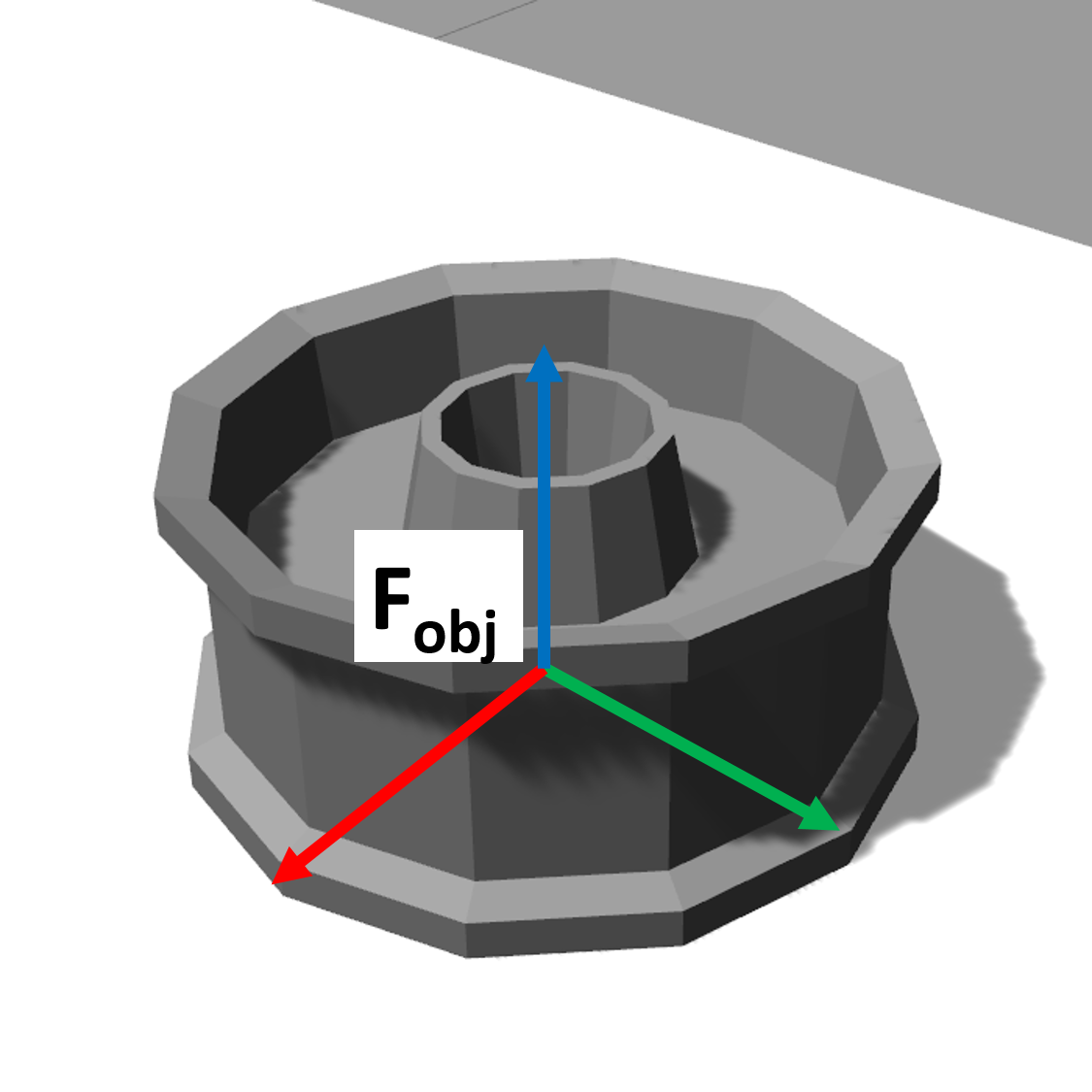}} &
\raisebox{-0.5\height}{\includegraphics[width=0.25\linewidth, keepaspectratio]{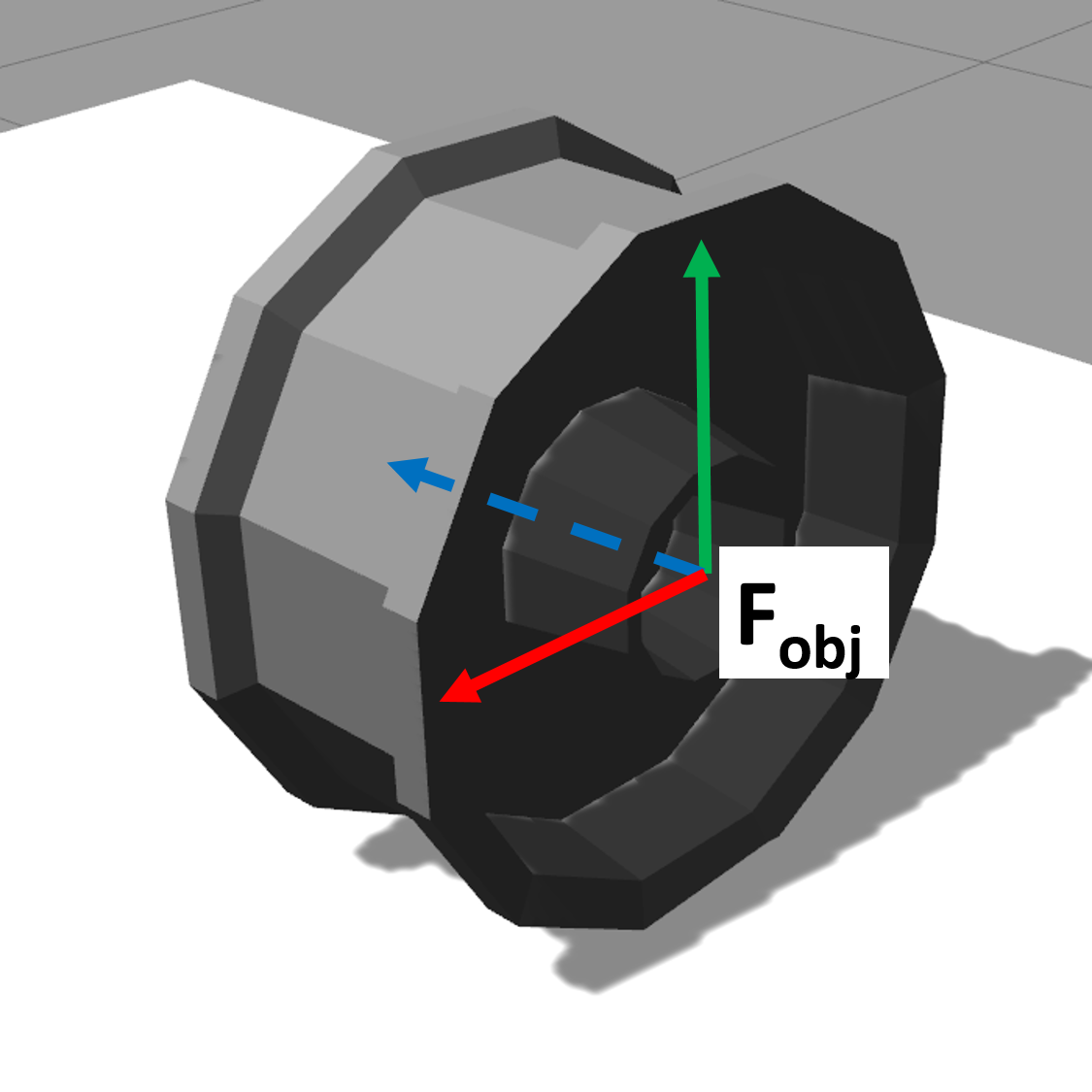}} &
\end{tabular}
\caption{The chosen objects and their frame $F_{obj}$ in their different stable positions.}
\label{fig:object_stable}
\end{figure}

Their 3D meshes used in the simulation in their different stable positions are visible on Fig. \ref{fig:object_stable}. Those objects were chosen for their relative complexity and diversity in terms of shapes.

\begin{figure}[htb]
\setlength{\tabcolsep}{0pt}
\renewcommand{\arraystretch}{0}
\centering
\begin{tabular}{c c c c c}
\includegraphics[width=0.19\linewidth, keepaspectratio]{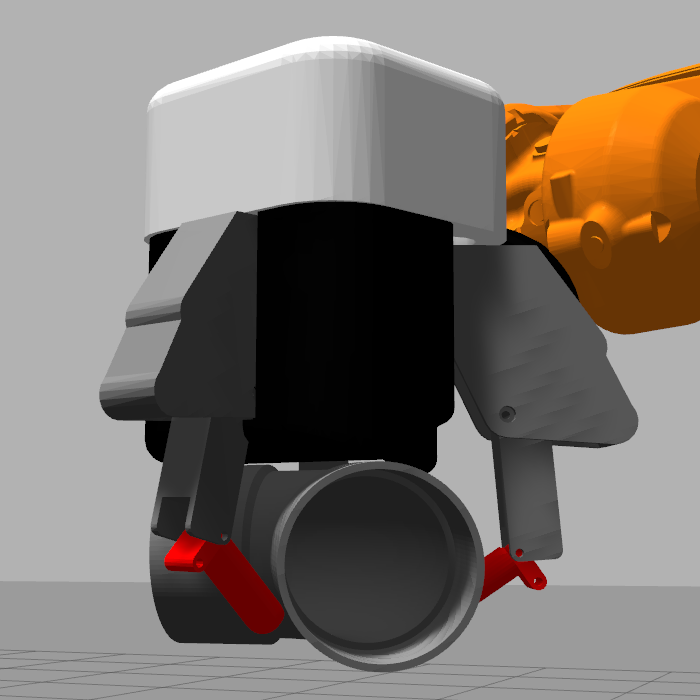} & \includegraphics[width=0.19\linewidth, keepaspectratio]{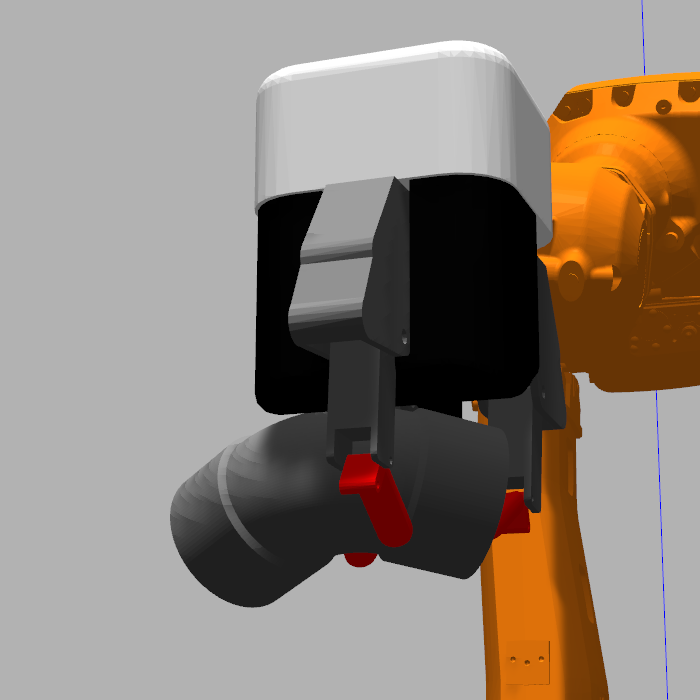} & \includegraphics[width=0.19\linewidth, keepaspectratio]{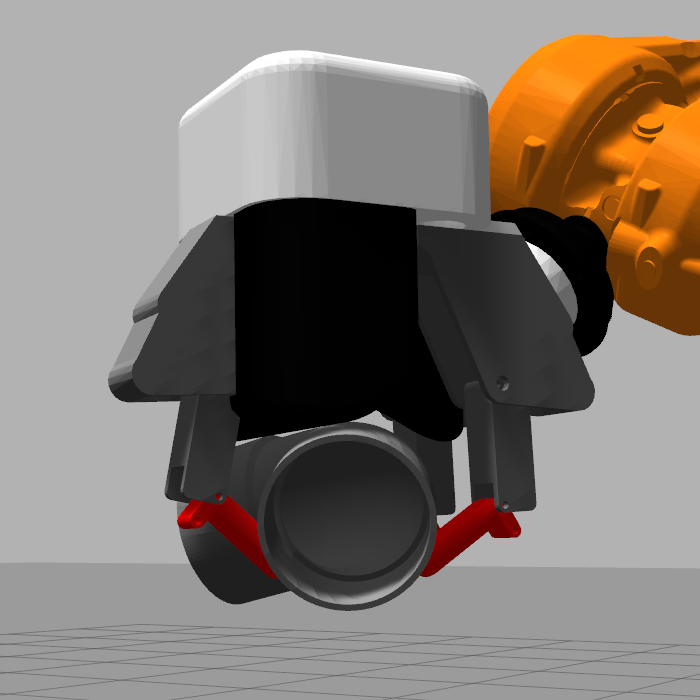} &
\includegraphics[width=0.19\linewidth, keepaspectratio]{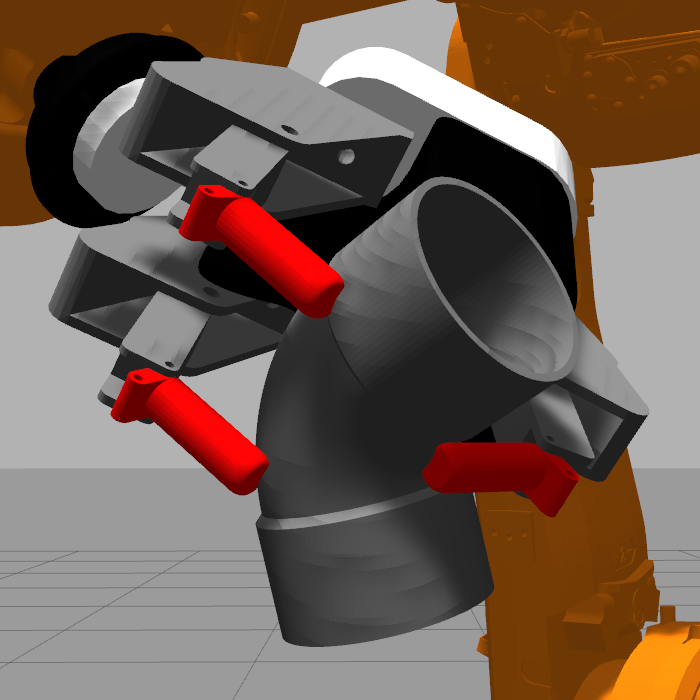} &
\includegraphics[width=0.19\linewidth, keepaspectratio]{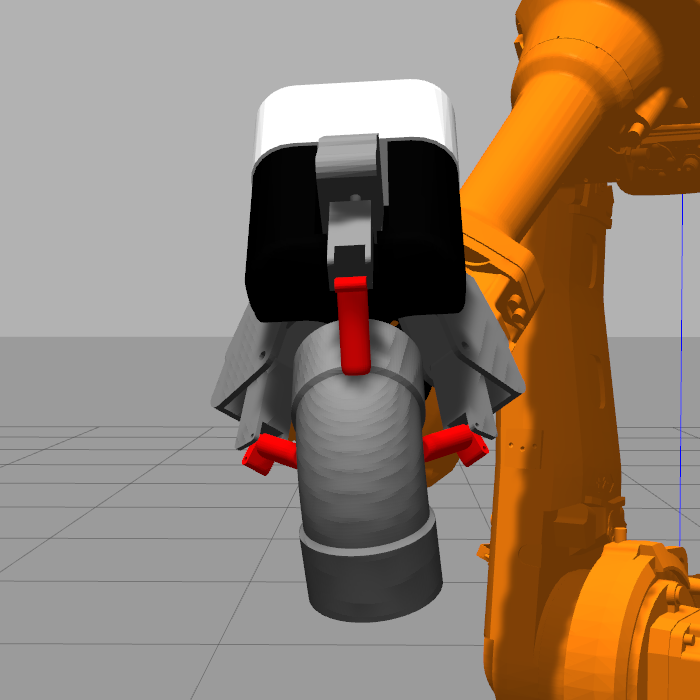}
\\
\includegraphics[width=0.19\linewidth, keepaspectratio]{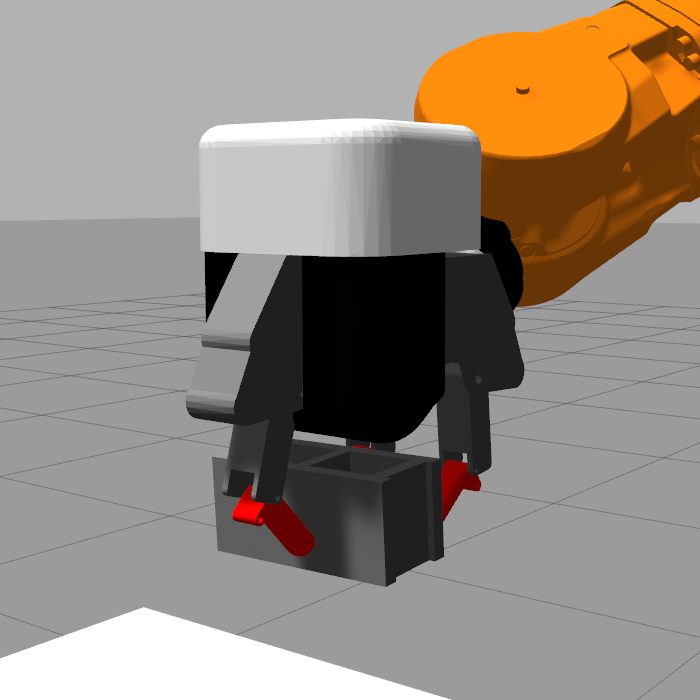} &
\includegraphics[width=0.19\linewidth, keepaspectratio]{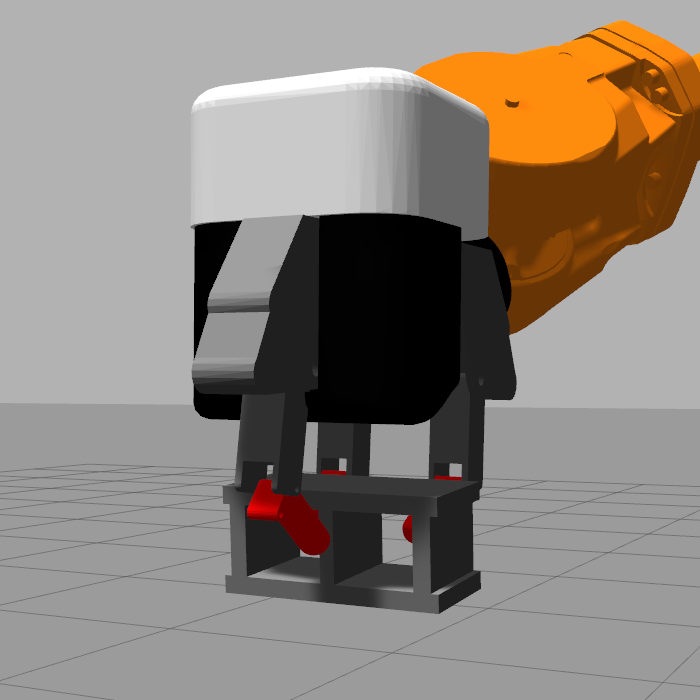} &
\includegraphics[width=0.19\linewidth, keepaspectratio]{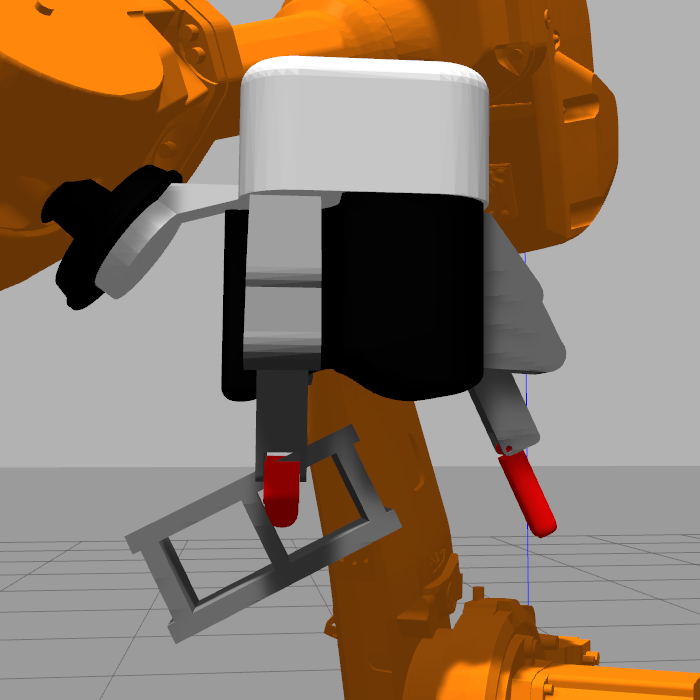} &
\includegraphics[width=0.19\linewidth, keepaspectratio]{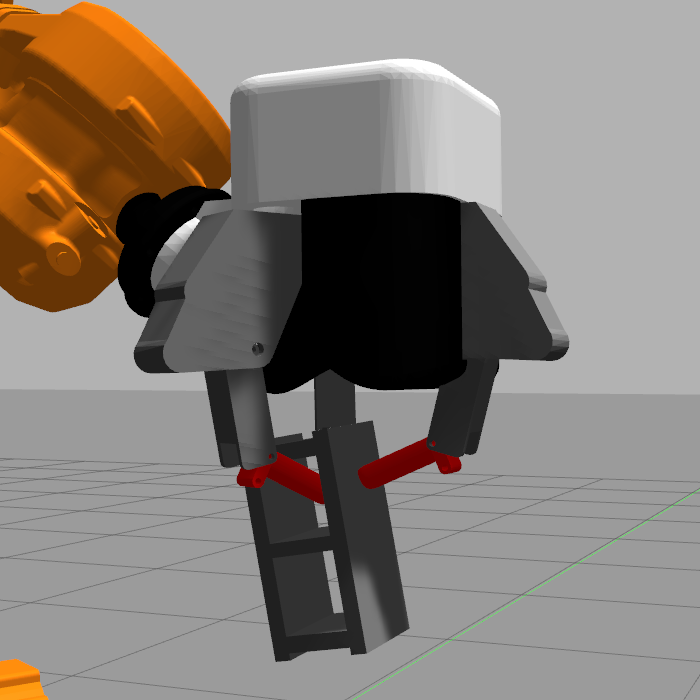} &
\includegraphics[width=0.19\linewidth, keepaspectratio]{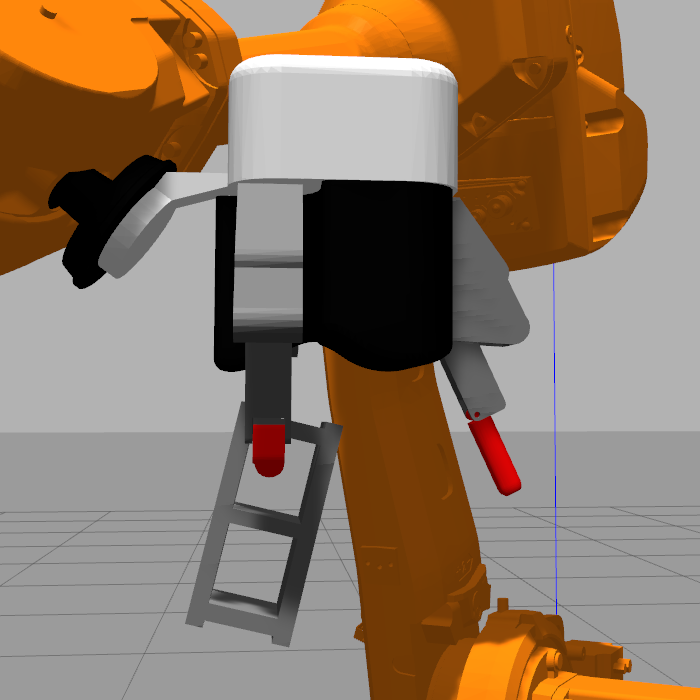}
\\
& \includegraphics[width=0.19\linewidth, keepaspectratio]{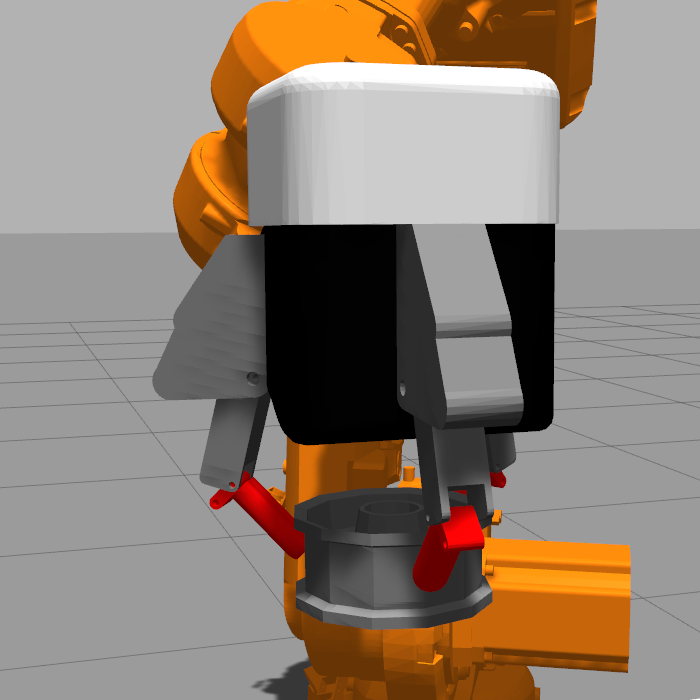} &
\includegraphics[width=0.19\linewidth, keepaspectratio]{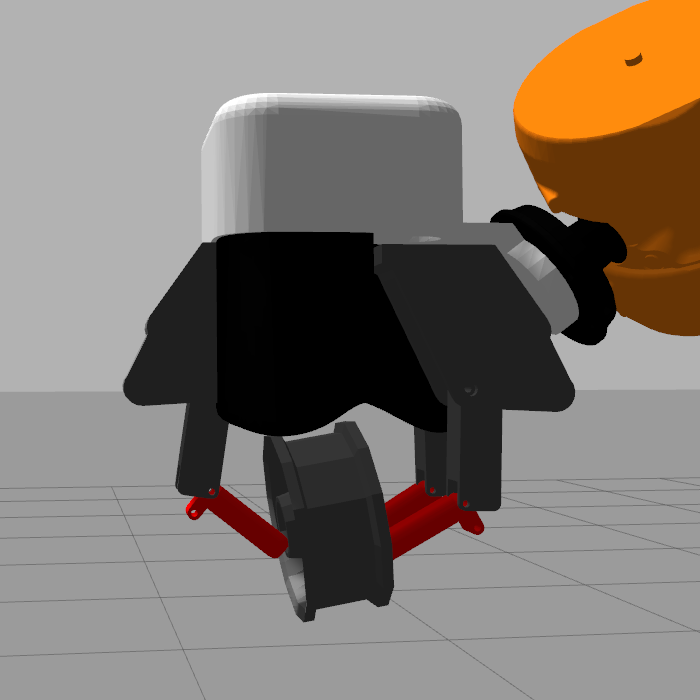} &
\includegraphics[width=0.19\linewidth, keepaspectratio]{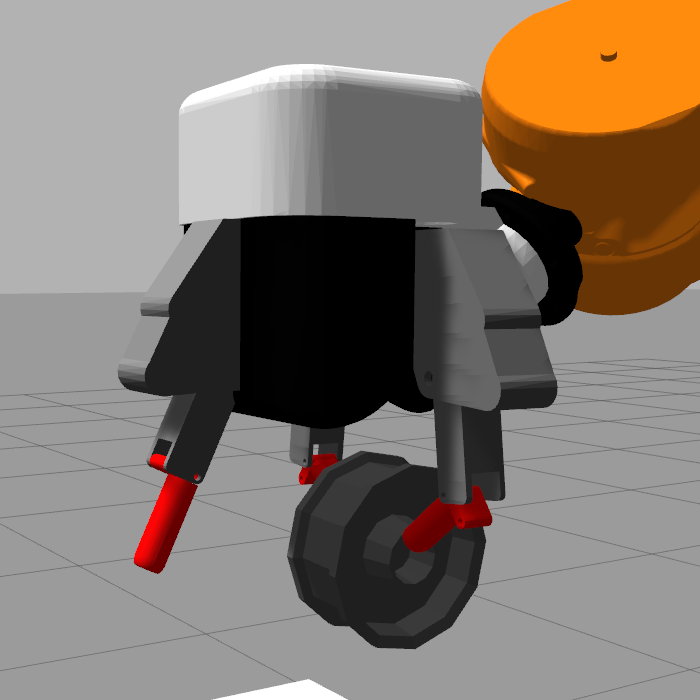} & 
\end{tabular}
\caption{Primitive grasp types for the three chosen objects. On the first row the grasp types for the bent pipe, on the second row for the cinder block, and on the third row for the pulley}
\label{fig:grasp_type}
\end{figure}

A set of primitive gripper configurations is determined for each of those objects for each of their stable position. These primitive gripper configurations can be sorted in different grasp types presented on Fig. \ref{fig:grasp_type}. For each of these grasp type, several variants are manually created.

For each object is gathered the following number of primitive grasps:
\begin{itemize}
\item bent pipe: 145 samples
\item cinder block: 141 samples
\item pulley: 118 samples
\end{itemize}

Around one hour is needed for a human operator to register the primitives for a given object.

The dataset stores the  eight parameters describing each primitive grasp along with the four parameters of the tabletop plane Cartesian equation in object frame $F_{obj}$. Indeed, many objects have different possible stable positions on the table. This is a critical information to avoid collisions with it. Some grasps may collide with the table in a given stable position, while being suitable for an other stable position. 

Expressing the grasp configuration in the object frame is still useful as it allows an invariance to a position change and to a rotation around a vertical axis.

\subsection{Human-initiated Grasp Generator VAE (HGG)}

The goal of the Human-initiated Grasp Generator VAE (HGG) is to infer the correlations existing between the parameters of different grasp primitives to learn a model of the grasp space. Such correlations exist, as primitive grasps are in the grasp space, and this space is a subset of the gripper configuration space. The HGG is able to use those correlations to map the grasp space in its latent space. This model can be used to generate efficiently new configurations that are likely to be in the grasp space. 

A distinct HGG is trained on the primitive grasp dataset for each object. Its inputs and outputs are shown in Fig. \ref{fig:archi} along with its global architecture. Before the training, the inputs and outputs data are normalized. This allows a faster training as the network does not have to scale its data by itself. For the gradient descent during the training, a Mean Square Error (MSE) is computed for each gripper parameter. Each of these errors is averaged on each batch. The global loss is computed as the sum of these averaged errors together with the KL divergence loss.

To make sure that the quaternion outputs by the decoder is a unit one, a custom activation function is used to normalize it on the output layer of the decoder.

\subsection{Latent Space Produced with Two Latent Variables}

\begin{figure}[htb]
\setlength{\tabcolsep}{0pt}
\renewcommand{\arraystretch}{0}
\centering
\begin{tabular}{c c c c c}
\includegraphics[width=0.19\linewidth, keepaspectratio]{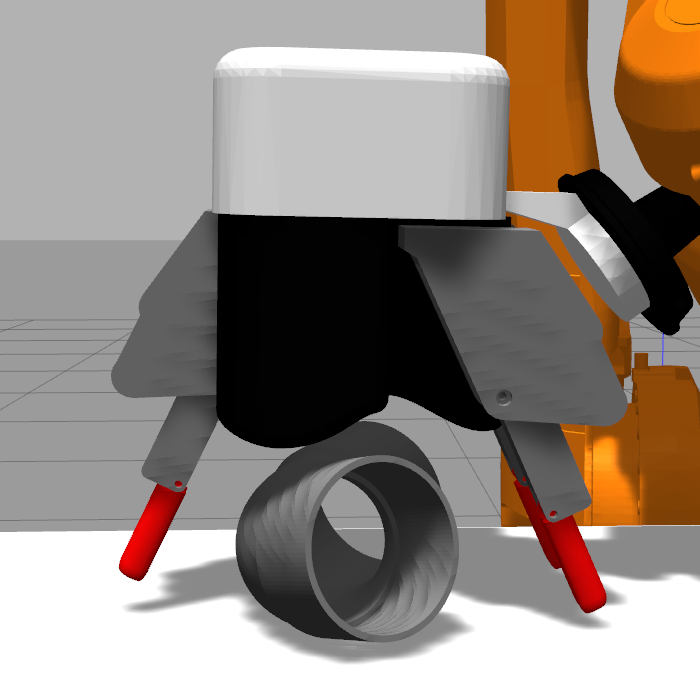} &
\includegraphics[width=0.19\linewidth, keepaspectratio]{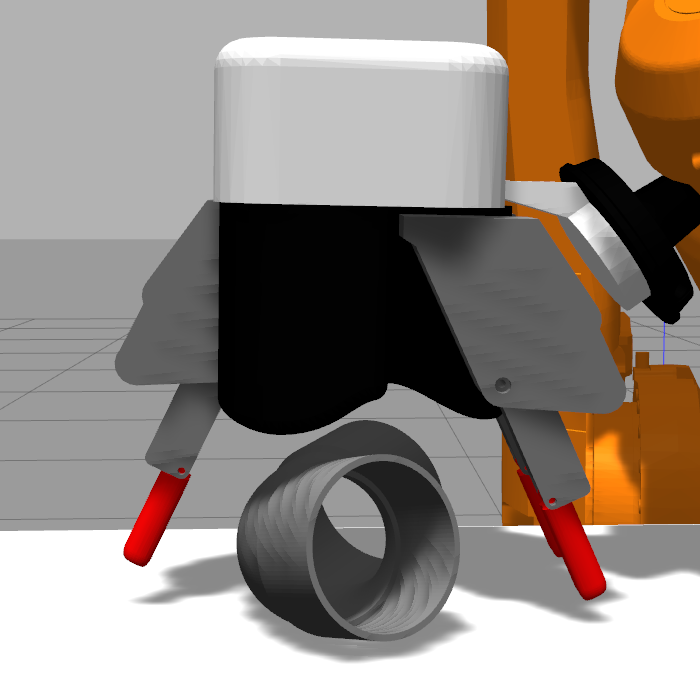} &
\includegraphics[width=0.19\linewidth, keepaspectratio]{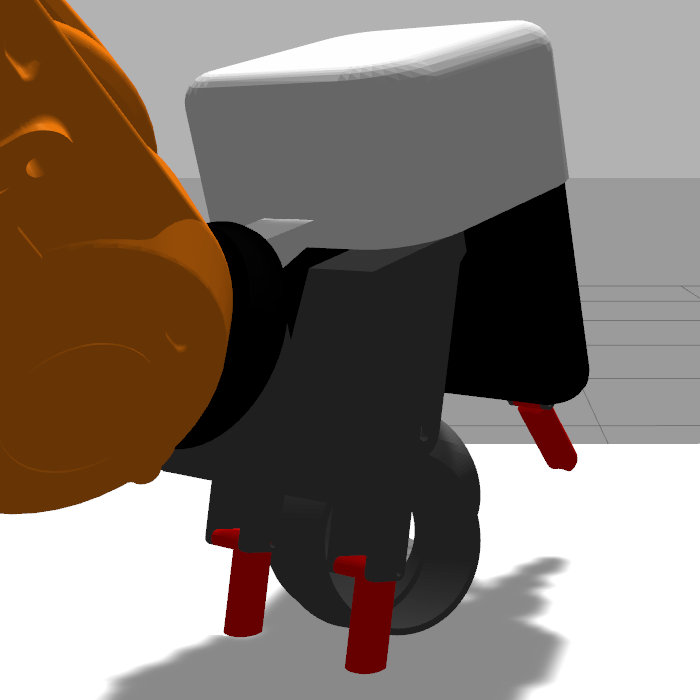} &
\includegraphics[width=0.19\linewidth, keepaspectratio]{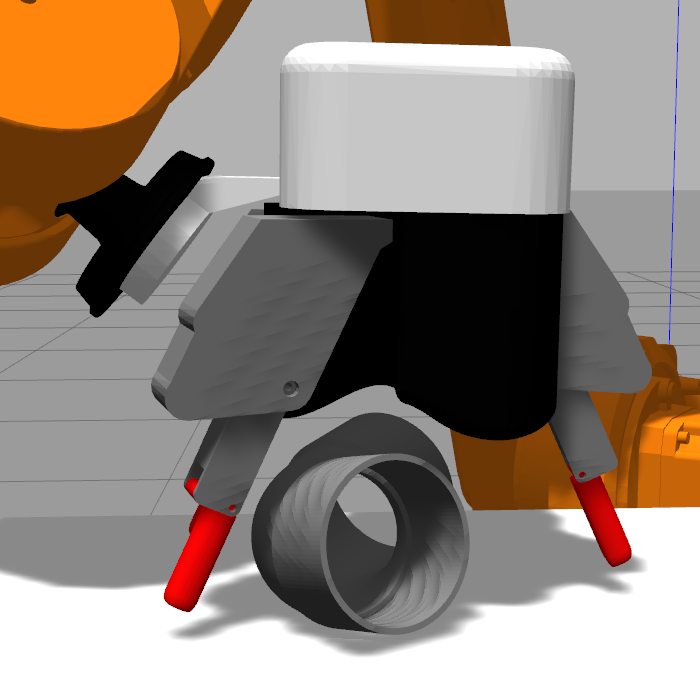} &
\includegraphics[width=0.19\linewidth, keepaspectratio]{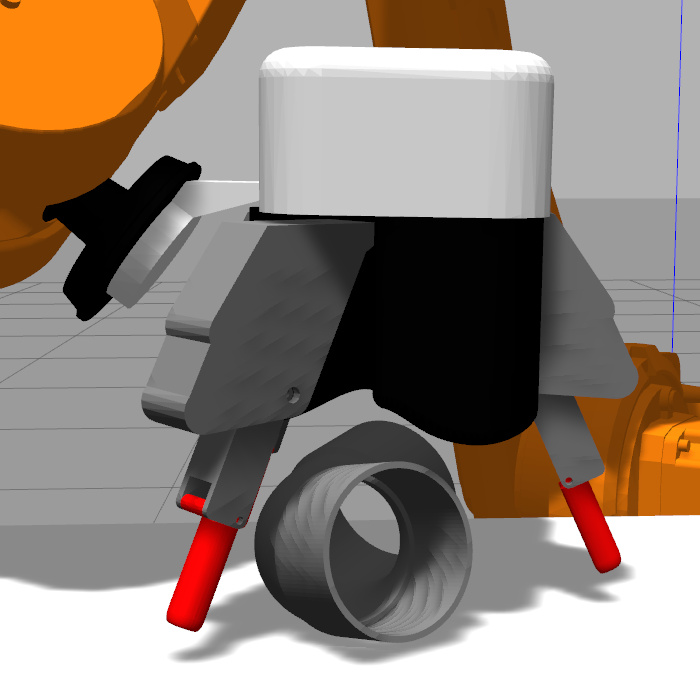}
\\
\includegraphics[width=0.19\linewidth, keepaspectratio]{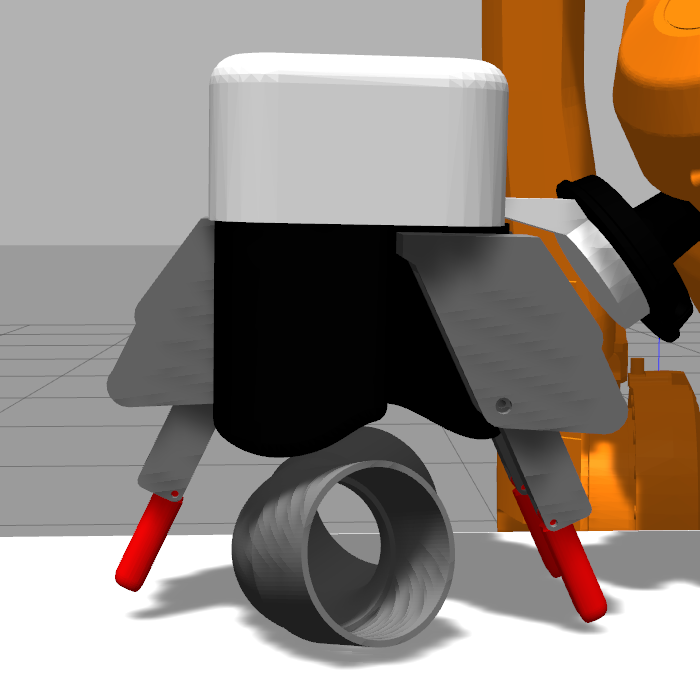} &
\includegraphics[width=0.19\linewidth, keepaspectratio]{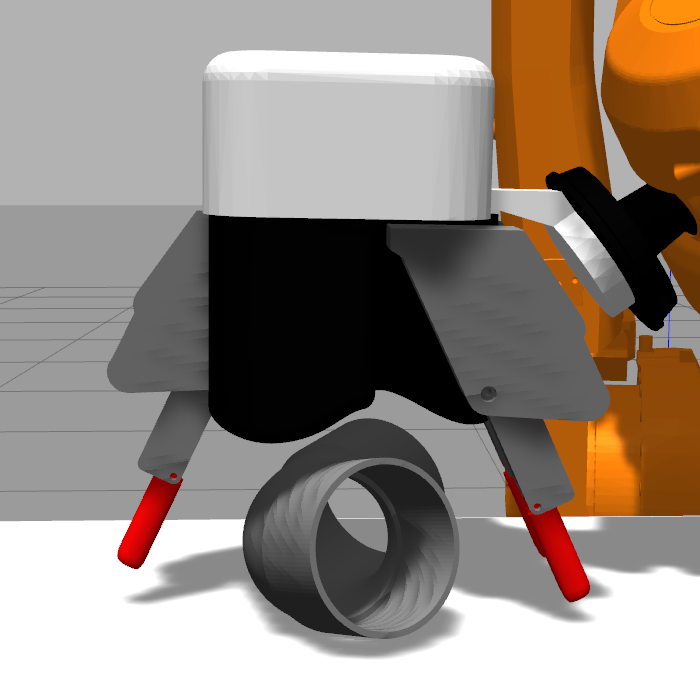} &
\includegraphics[width=0.19\linewidth, keepaspectratio]{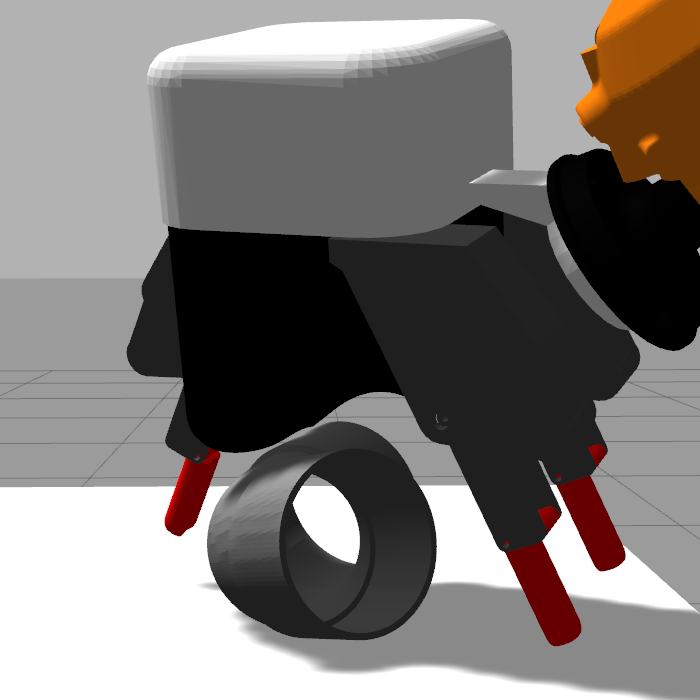} &
\includegraphics[width=0.19\linewidth, keepaspectratio]{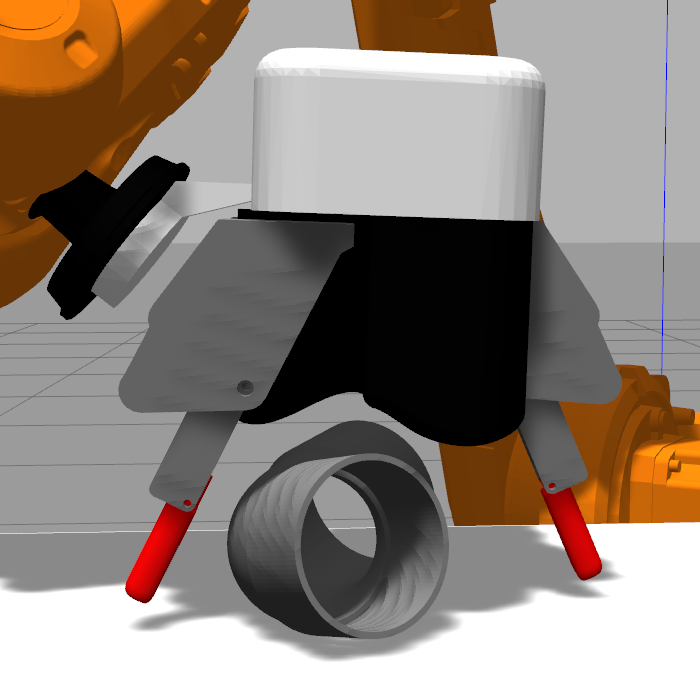} &
\includegraphics[width=0.19\linewidth, keepaspectratio]{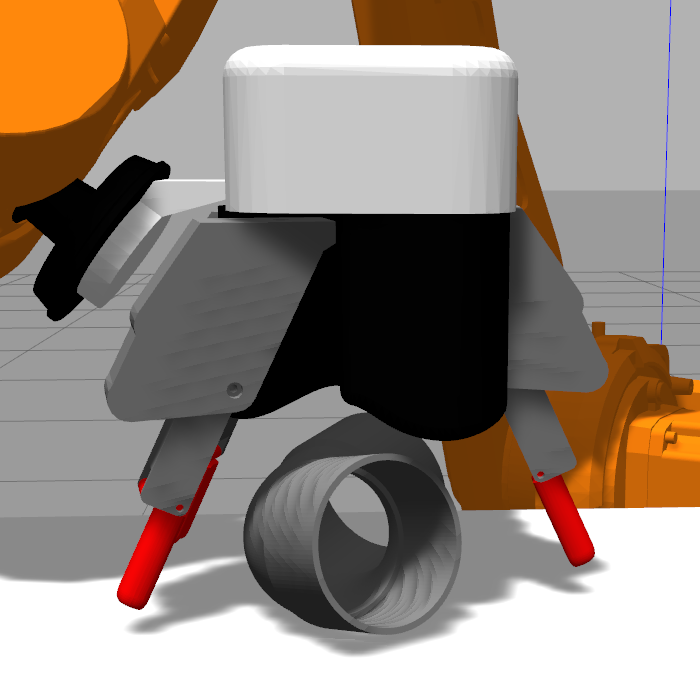}
\\
\includegraphics[width=0.19\linewidth, keepaspectratio]{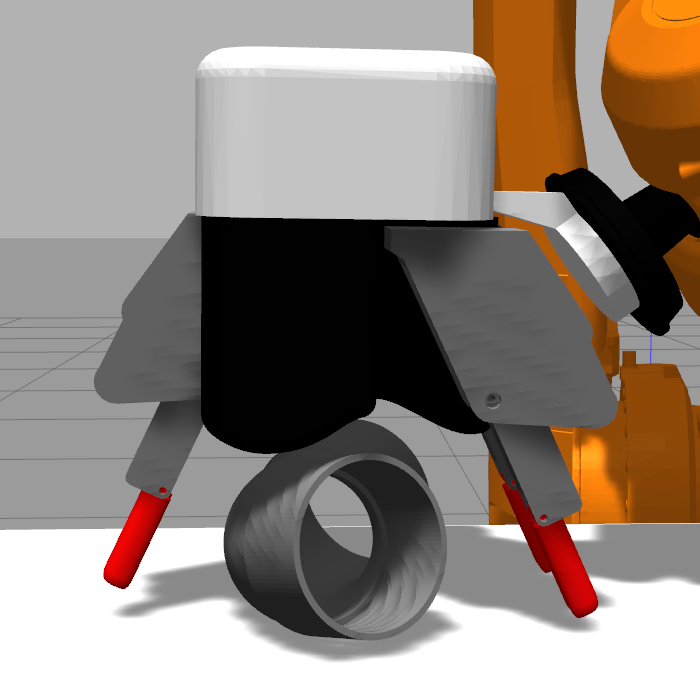} &
\includegraphics[width=0.19\linewidth, keepaspectratio]{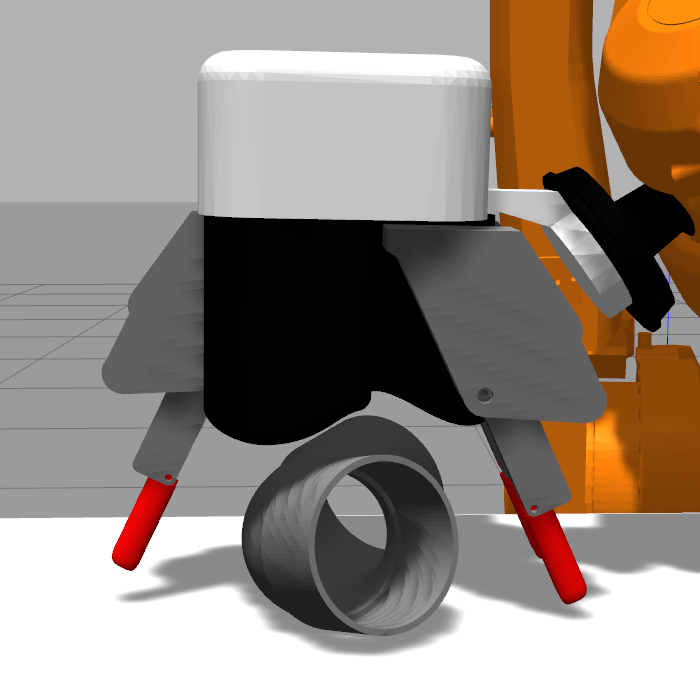} &
\includegraphics[width=0.19\linewidth, keepaspectratio]{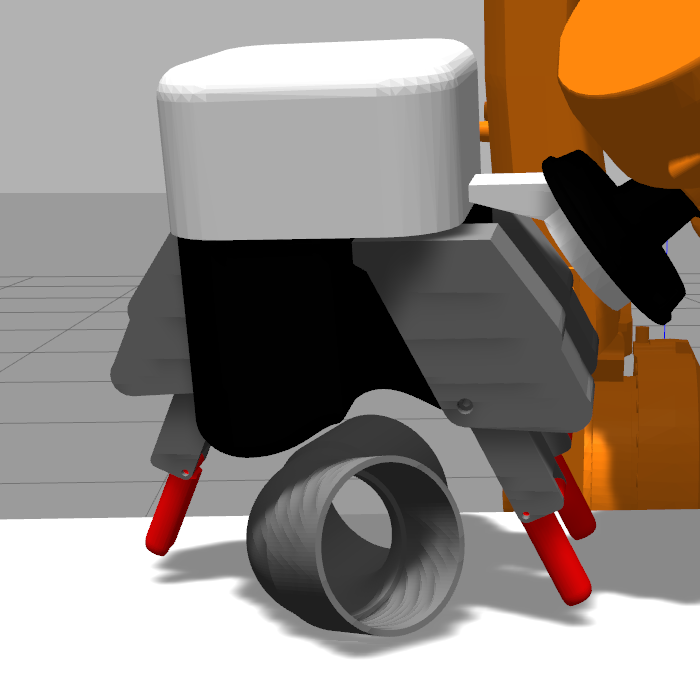} &
\includegraphics[width=0.19\linewidth, keepaspectratio]{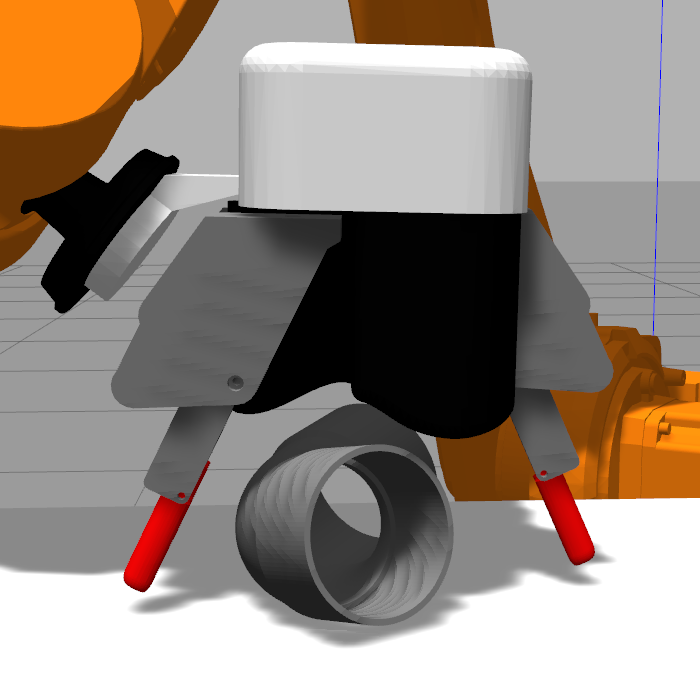} &
\includegraphics[width=0.19\linewidth, keepaspectratio]{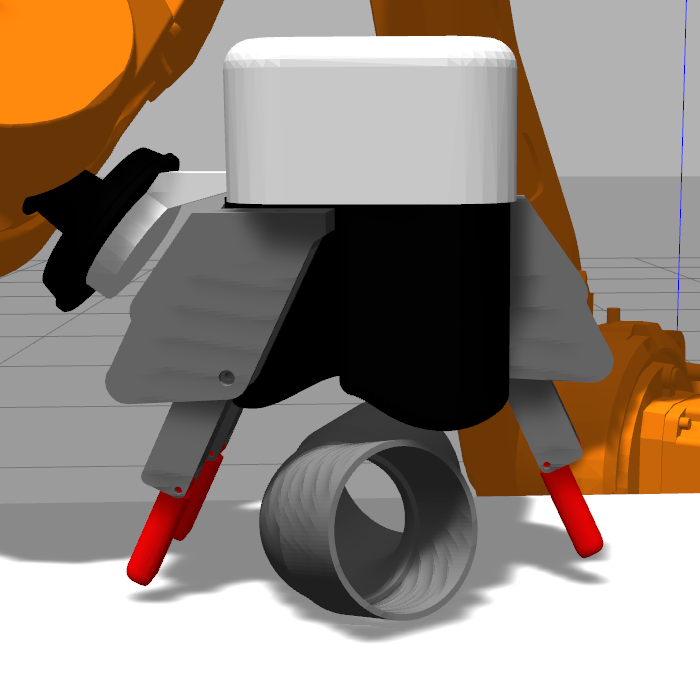}
\\
\includegraphics[width=0.19\linewidth, keepaspectratio]{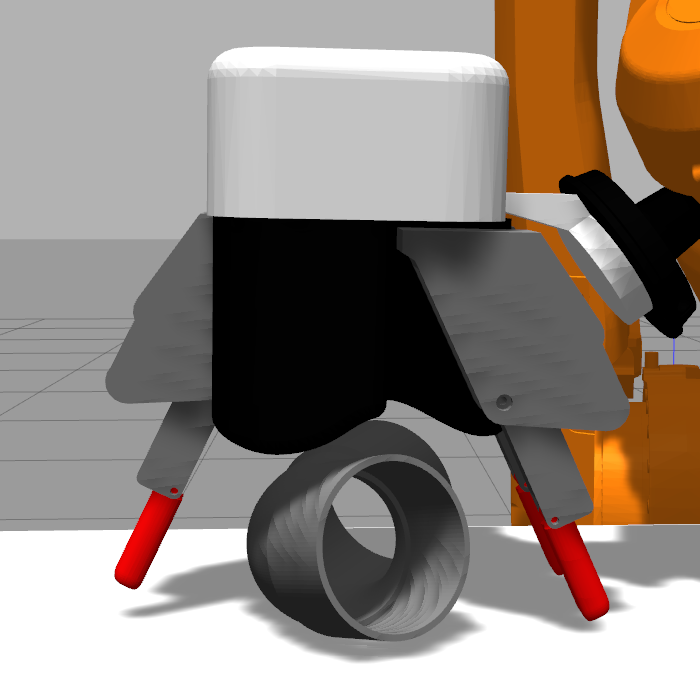} &
\includegraphics[width=0.19\linewidth, keepaspectratio]{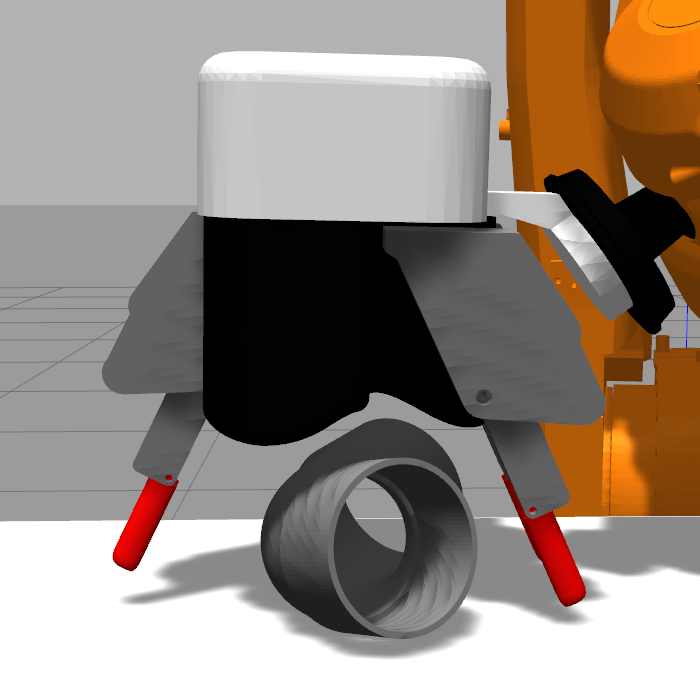} &
\includegraphics[width=0.19\linewidth, keepaspectratio]{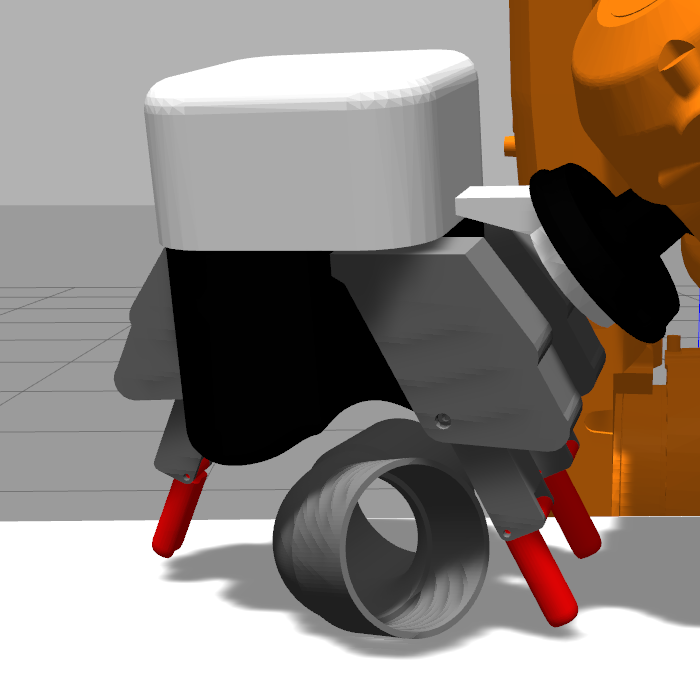} &
\includegraphics[width=0.19\linewidth, keepaspectratio]{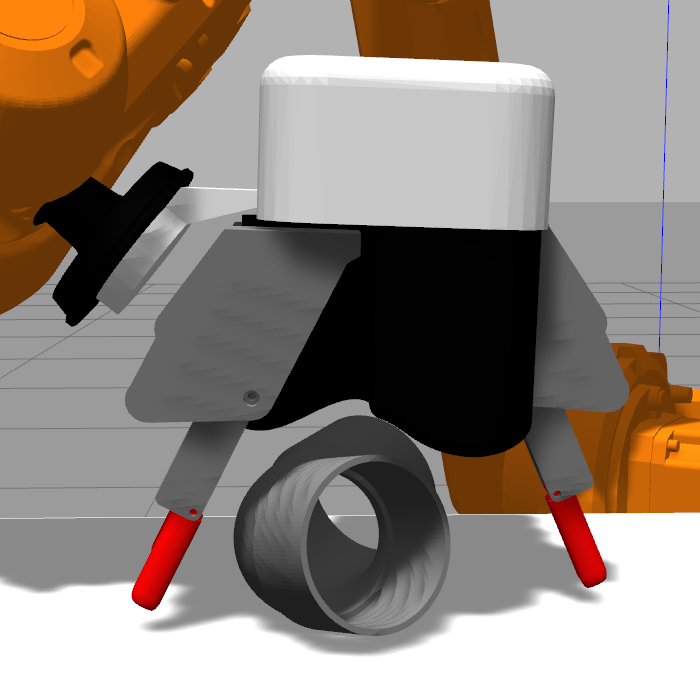} &
\includegraphics[width=0.19\linewidth, keepaspectratio]{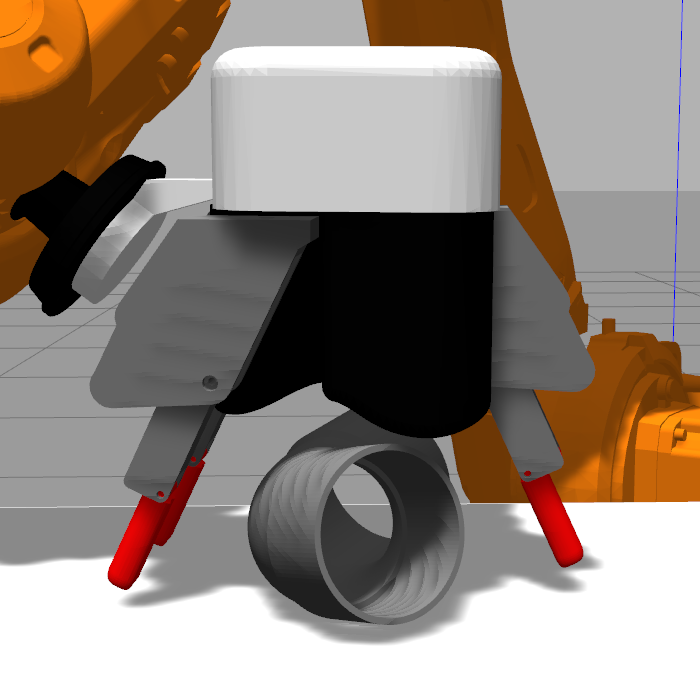}
\\
\includegraphics[width=0.19\linewidth, keepaspectratio]{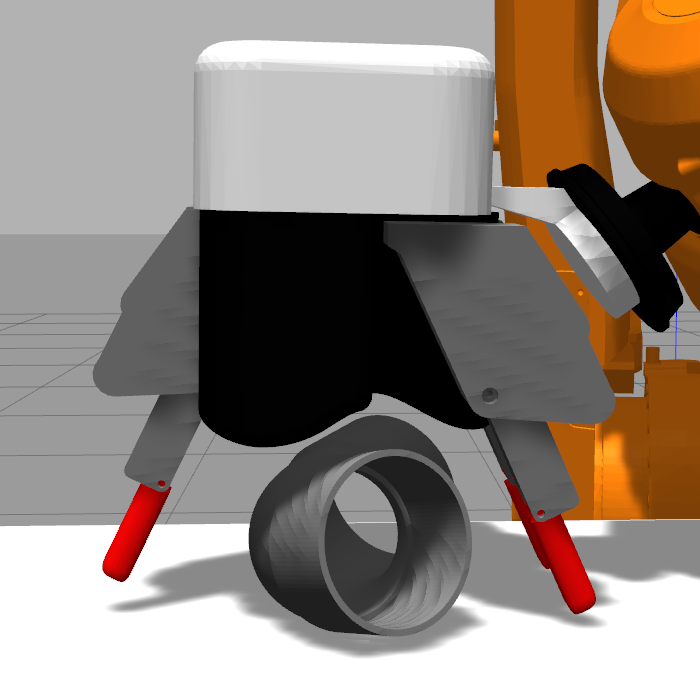} &
\includegraphics[width=0.19\linewidth, keepaspectratio]{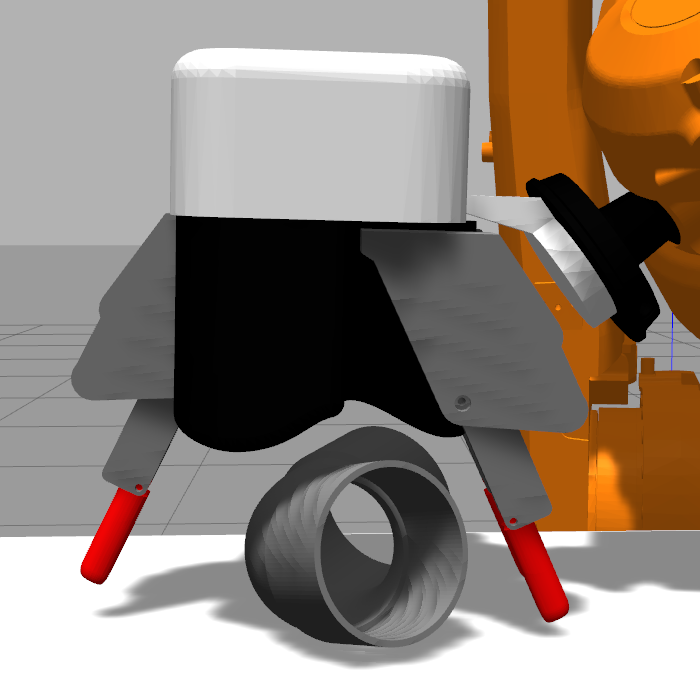} &
\includegraphics[width=0.19\linewidth, keepaspectratio]{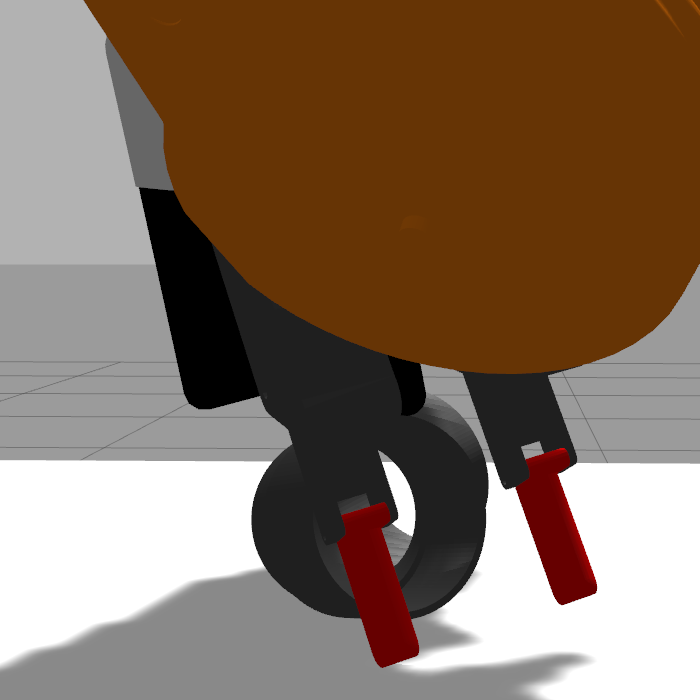} &
\includegraphics[width=0.19\linewidth, keepaspectratio]{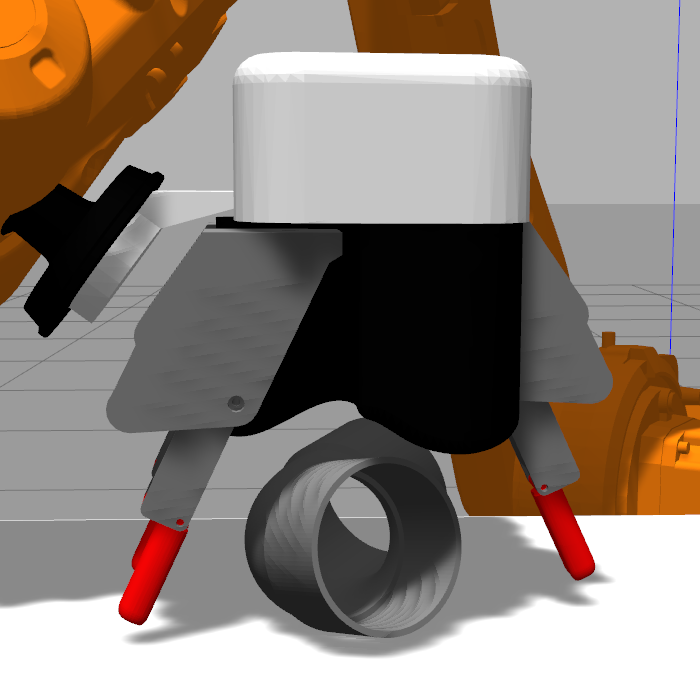} &
\includegraphics[width=0.19\linewidth, keepaspectratio]{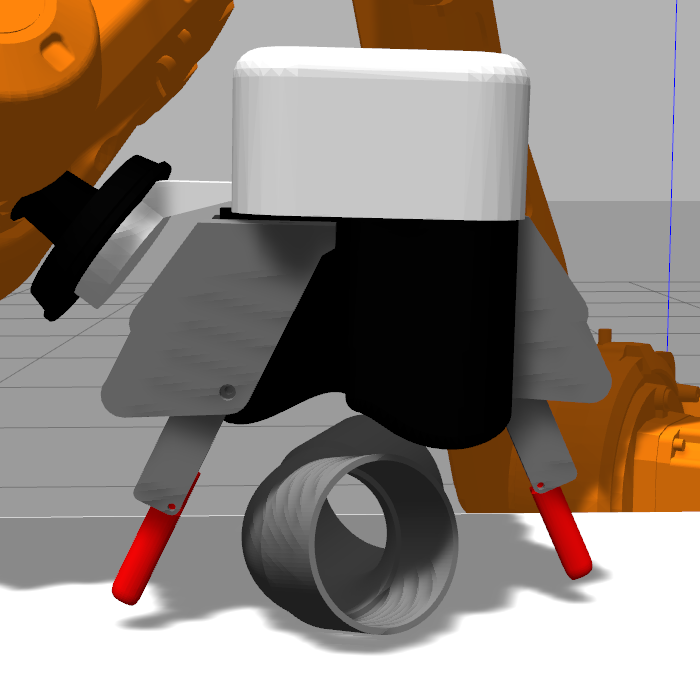}
\end{tabular}
\caption{Gripper configurations generated when visiting a two dimensional latent space. The central image is the point (0, 0) in latent space. The inner and outer image rings around it correspond to points evenly distributed on circles of diameter respectively 0.5 and 1 in latent space. Here, translations along the image plan normal are not visible, which explains some visually almost identical configurations.}
\label{fig:grasp_model}
\end{figure}

The HGG learns to model the grasp space in its latent space. By sampling values in it, one can generate new gripper configurations that are likely to belong to the grasp space. On Fig.  \ref{fig:grasp_model} is shown the obtained gripper configurations when visiting a two dimensional latent space for one stable position of the bent pipe. As some configurations may not lead to successful grasps, or may be in collision with the object or the environment, only pre-grasp configuration are shown (that is before closing the fingers), with collisions disabled.

For this stable position, it appears that the latent variable displayed on the horizontal axis on Fig \ref{fig:grasp_model} encodes mainly the direction from which the bent pipe will be grasped: the two rearrangeable fingers on the concave side or on the convex side. The other latent variable seems to encode mainly translations. On the top-right corner appears a configuration corresponding to the first bent pipe primitive grasp type (shown on Fig. \ref{fig:grasp_type}). The rest of the right side configurations correspond to the third grasp type, and the left side configurations to the second grasp type.

%% file: scheme_architecture.tex
\begin{tikzpicture}[every node/.append style={font=\footnotesize}]

\tikzstyle{inputNode}=[draw, fill=blue!20]
\tikzstyle{hiddenNode}=[draw, fill=black!30!green!20]
\tikzstyle{outputNode}=[draw, fill=red!20]
\tikzstyle{inputArrow}=[->, >=latex, blue]
\tikzstyle{hiddenArrow}=[->, >=latex, black!30!green]
\tikzstyle{outputArrow}=[->, >=latex, red]

\draw[rounded corners] (-0.6,2.2) rectangle (11.2,-1.75);
\node[] at (5.3,1.9) {Encoder};
\node[inputNode, minimum height=1.1cm, minimum width=2.05cm, align=center] (conf-enc) at (0.55, 0.7){gripper \\ configuration, \\ $x$, $y$, $z$, $\theta$, \\ $q_x$, $q_y$, $q_z$, $q_w$};
\node[inputNode, minimum height=1.1cm, minimum width=2.05cm, align=center] (table-enc) at (0.55, -1.1) {tabletop \\ equation, \\ $a$, $b$, $c$, $d$};

\node[minimum height=3.6cm, minimum width=1.9cm](dummy-node-preenc) at (3., 0.25){};
\node[hiddenNode, minimum height=0.9cm, minimum width=1.9cm, align=center](prepos-enc) at (3.,1.6) {position \\ input NN};
\node[hiddenNode, minimum height=0.9cm, minimum width=1.9cm, align=center](preori-enc) at (3.,0.7) {orientation \\ input NN};
\node[hiddenNode, minimum height=0.9cm, minimum width=1.9cm, align=center](presp-enc) at (3.,-0.2) {spread angle \\ input NN};
\node[hiddenNode, minimum height=0.9cm, minimum width=1.9cm, align=center](pretable-enc) at (3.,-1.1) {tabletop \\ input NN};

\node[hiddenNode, minimum height=2.2cm, minimum width=3cm] (main-enc) at (6.3,0.25) {main encoding NN};

\node[minimum height=1.4cm, minimum width=0.375cm](dummy-node-output) at (10.9, 0.25){};
\node[outputNode, minimum height=0.7cm] (out1-enc) at (10.9,-0.1) {$n$};
\node[outputNode, minimum height=0.7cm] (out2-enc) at (10.9,0.60) {$n$};

\node[draw, minimum width=1.9cm] (mean) at (13.8,-0.1) {means};
\node[draw, minimum width=1.9cm] (variance) at (13.8,0.60) {log-variances};

\draw[rounded corners] (-0.6,-2.1) rectangle (11.2,-5.7);
\node[font=\footnotesize] at (5.3,-2.4) {Decoder};
\node[inputNode, minimum width=1.5cm, minimum height=1.1cm, align=center] (in1-dec) at (0.5,-4.95) {tabletop \\ equation, \\ $a$, $b$, $c$, $d$};
\node[inputNode, minimum width=1.5cm, minimum height=1.1cm, align=center] (in2-dec) at (0.5,-3.55) {sampling \\ result, \\ $l_1, ..., l_n$};

\node[hiddenNode, minimum height=1.1cm, minimum width=1.5cm, align=center](pretable-dec) at (3.,-4.95) {tabletop \\ input NN};

\node[hiddenNode, minimum height=2.2cm, minimum width=3cm] (main-dec) at (6.3,-4.25) {main decoding NN};

\node[hiddenNode, minimum height=0.9cm, minimum width=1.9cm, align=center](postpos-dec) at (9.4,-3.35) {position \\ output NN};
\node[hiddenNode, minimum height=0.9cm, minimum width=1.9cm, align=center](postori-dec) at (9.4,-4.25) {orientation \\ output NN};
\node[hiddenNode, minimum height=0.9cm, minimum width=1.9cm, align=center](postsp-dec) at (9.4,-5.15) {spread angle \\ output NN};

\node[outputNode, minimum height=0.8cm] (outpos-dec) at (10.9,-3.35) {3};
\node[outputNode, minimum height=0.8cm] (outori-dec) at (10.9,-4.25) {4};
\node[outputNode, minimum height=0.8cm] (outspread-dec) at (10.9,-5.15) {1};

\draw[rounded corners] (12.7, -2.15) rectangle (14.9, -5.70);
\node[align=center] (config) at (13.8,-2.55) {gripper \\ configuration};
\node[draw, align=center, minimum height=0.9cm, minimum width=2.0cm] (pos) at (13.8,-3.35) {position \\ $x$, $y$, $z$};
\node[draw, align=center, minimum height=0.9cm, minimum width=2.0cm] (ori) at (13.8,-4.25) {orientation \\ $q_x$, $q_y$, $q_z$, $q_w$};
\node[draw, align=center, minimum height=0.9cm, minimum width=2.0cm] (spread) at (13.8,-5.15) {spread \\ angle $\theta$};

\draw[inputArrow, very thick] (-1, 0.7) -- (conf-enc.west);
\draw[inputArrow, very thick] (-1, -1.1) -- (table-enc.west);
\draw[inputArrow] (conf-enc.east) -- (prepos-enc.west);
\draw[inputArrow] (conf-enc.east) -- (preori-enc.west);
\draw[inputArrow] (conf-enc.east) -- (presp-enc.west);
\draw[inputArrow] (table-enc.east) -- (pretable-enc.west);
\draw[hiddenArrow] (dummy-node-preenc.east) -- (main-enc.west);
\draw[hiddenArrow] (main-enc.east) -- (dummy-node-output.west);
\draw[outputArrow] (out1-enc) -- (mean);
\draw[outputArrow] (out2-enc) -- (variance);

\draw[inputArrow, very thick] (-1, -4.95) -- (in1-dec);
\draw[inputArrow, very thick] (variance.east) -| (14.9, -0.7) |- (-1, -1.925) -| (-1,-3) |- (in2-dec.west);
\draw[inputArrow, very thick] (mean.east) -| (14.9, -0.7) |- (-1, -1.925) -| (-1,-3) |- (in2-dec.west);
\draw[inputArrow] (in2-dec.east) -- (main-dec.west);
\draw[inputArrow] (in1-dec.east) -- (pretable-dec.west);
\draw[hiddenArrow] (pretable-dec.east) -- (main-dec.west);
\draw[hiddenArrow] (main-dec.east) -- (postori-dec.west);
\draw[hiddenArrow] (postpos-dec.east) -- (outpos-dec.west);
\draw[hiddenArrow] (postori-dec.east) -- (outori-dec.west);
\draw[hiddenArrow] (postsp-dec.east) -- (outspread-dec.west);

\draw[outputArrow] (outpos-dec.east) -- (pos.west) node[midway, above, black, align=center]{sigmoid};
\draw[outputArrow] (outori-dec.east) -- (ori.west) node[midway, black, align=center]{quaternion \\ normalizer};
\draw[outputArrow] (outspread-dec.east) -- (spread.west) node[midway, below, black, align=center]{sigmoid};
\end{tikzpicture}

%% file: vae_tuning.tex
The HGG has three main hyperparameters that can be tuned to improve the learnt grasp space model:
\begin{itemize}
\item the network size;
\item the latent space dimension;
\item the KL divergence loss component coefficient \citep{Higgins_beta_2017}.
\end{itemize}
Several indicators can be monitored to assess the effect of those hyperparameters on the performances of the HGG:
\begin{itemize}
\item the reconstruction error;
\item the KL divergence loss component value;
\item the number of used latent variables, that is the number of latent variables with a high KL divergence;
\item the share of generated successful grasps.
\end{itemize}

Various learning trials were conducted with different hyperparameters combinations. A summary of the effects of the hyperparameters is given in the following subsections.

\subsection{Trade-Off Between Reconstruction and Regularity}

One of the distinctive features of a VAE is that its loss function combines a reconstruction cost and a regularization cost, the KL divergence cost. This leads the training process to converge to a trade-off between reconstruction and regularity. The reconstruction is the ability to accurately reproduce on the output the input data. The regularity is the fact that the input data are homogeneously distributed in each latent variable (here, following a normal distribution), and that latent variables are disentangled. A side effect of those constraints is that the network is pushed to use as few latent variables as possible to represent the data. For the HGG, both terms are important: a good reconstruction is needed as it allows to capture faithfully all the primitive data variability, and a good regularity is also needed as it reduces the data distribution sparsity, and thus the risk of generating inconsistent configurations. 

In \cite{Higgins_beta_2017}, the authors introduced a coefficient on the KL divergence term that allows to adjust this trade-off. Increasing this coefficient will put higher priority on the KL divergence term, and thus increases the regularity at the expense of the reconstruction. A too high coefficient on this term can push the network to ignore the data variability along a given axis to homogenize the data in its latent variables and disentangle them, leading to poor reconstruction. Conversely, a too low coefficient will allow a very accurate reconstruction, but the latent variables will be more entangled and the data distribution in them will be sparser.

The optimal value of the coefficient depends mainly on the latent space dimension. In \cite{Higgins_beta_2017}, they recommend a value greater than 1, but they use the VAE for image generation, which involve generally a latent space of greater dimension than for the presented use case. For the HGG, a value below 1 is mandatory to keep an acceptable reconstruction loss.

\subsection{Network Size}

To increase reconstruction with less impact on regularity than the KL coefficient, one can increase the network size, that is the number of neurons in the different layers, or the number of layers. Indeed, it increases the number of network trainable parameters, and thus the complexity of the functions that it is able to approximate. However, it increases the computational cost of both the learning phase and the inference phase, and the memory required to store model. Moreover, the more parameters the network has, the more training data it needs for a proper training. For the HGG use case, there are very few training data, which limit the size of the network. 

For the architecture presented in Fig. \ref{fig:archi}, a good trade-off between computation cost and general performances is around $30\,000$ parameters for the whole VAE, that is both encoder and decoder. Indeed, the reconstruction starts to improve less significantly beyond this threshold.

\subsection{Grasp Space Dimension \& Latent Space Dimension}

The grasp space is a subset of the gripper configuration space (see subsection \ref{subsec:prob_statement}). Thus, it has at most 7 dimensions, but its true size is a priori unknown. As the goal is to map the grasp space in the HGG latent space, it is important that the number of latent variables used by the HGG among the available ones is at least equal to the grasp space dimension. Otherwise, there will be information loss due to the compression caused by the projection of the grasp space into a smaller space. Although conservative, it is sub-optimal to let the HGG find by itself the required number of latent variables needed to map the grasp space, by letting the latent space dimension be equal to the one of the gripper configuration space. Indeed, even if the KL divergence term in the cost function will push the network to use as few latent variables as possible, increasing the number of available latent variables increases the chances to converge toward a cost function local minimum where the network uses more latent variable than needed. As the used latent space dimension is greater, the data distribution inside it is also sparser, which have the same effect as a too low coefficient on the KL divergence.

Thus, it is useful to know an approximation of the dimension of the grasp space. Here, a dimensional analysis tool has been chosen, the kernel-PCA \citep{Scholkopf_kernel_1999}. Kernel-PCA is an extension of the PCA to non-linear relations. Indeed, the grasp space is probably a submanifold of the gripper configuration space, which probably involves non-linear relations between the parameters of this space.

The kernel-PCA implemented in scikit-learn is used \citep{scikit-learn}. The algorithm is run for each object, taking as input the list of gripper configurations in the primitive grasp dataset. The grasp space dimension is determined as the number of eigenvectors of the centered kernel matrix needed to retrieve 90\% of the information, by looking at their eigenvalues. Indeed, it means that the kernel-PCA can explain 90\% of the data variability with the given number of eigenvectors. The outputted result depend on the object: 3 dimensions for the pulley, and 4 dimensions for the bent pipe and the cinder block.

Therefore, this can serve as an upper bound for the optimal latent space dimension. Indeed, the HGG has a supplementary information: the tabletop Cartesian equation, and may use it to learn a more compact representation than the one found by the kernel-PCA.

\subsection{Overview}

In table \ref{fig:summary} is summarized the influence of the three hyperparameters on the chosen indicators. For this evaluation, trials were conducted with hyperparameter combinations among the following ranges:
\begin{itemize}
\item network size between $12\,000$ and $30\,000$ parameters;
\item latent space dimension between 2 and 6;
\item KL divergence coefficient between $0.0002$ and $0.01$.
\end{itemize}

\begin{table}[htb]
\centering
\caption{Spearman correlation coefficients between the hyperparameters and the indicators.}
\input{summary_table.tex}
\label{fig:summary}
\end{table}

\begin{table}[htb]
\centering
\caption{performances of the HGG for the selected hyperparameters: $30\,000$ network parameters, 3 latent variables (that are all used), and a KL divergence coefficient of $0.0005$. The mean errors are measured on the training data.}
\input{hgg_perf.tex}
\label{fig:hgg_perf}
\end{table}

In table \ref{fig:hgg_perf} are shown the performances corresponding to the set of hyperparameters achieving the best trade-off between the reconstruction and the proportion of generated successful grasps. To avoid arm kinematic reachability issues, as gripper configurations are in object frame, each generated configuration is tested for different object orientations relative to the robot. The main cases of failing grasps are found when transitioning between different grasp types, and with the fifth grasp type of the bent pipe (Fig. \ref{fig:grasp_type}, top right) where one of the bottom finger can collide with the table in the pre-grasp phase for some gripper orientation variations.

%% file: summary_table.tex
\footnotesize
\begin{tabular}{ >{\raggedright\arraybackslash}m{2.6cm} >{\centering\arraybackslash}m{1.5cm} >{\centering\arraybackslash}m{1.5cm} >{\centering\arraybackslash}m{1.5cm} }
& latent space dimension & KL divergence coefficient & network size \\
\toprule
number of used latent variables & $0.12$ & $-0.26$ & $0.55$ \\
\hline
reconstruction error & $-0.18$ & $0.18$ & $-0.21$ \\
\hline
KL divergence & $-0.75$ & $-0.62$ & $0.09$ \\
\hline
proportion of generated successful grasps & $-0.39$ & $0.25$ & $0.39$ \\
\bottomrule
\end{tabular}

%% file: hgg_perf.tex
\footnotesize
\begin{tabular}{ >{\raggedright\arraybackslash}m{1.6cm} >{\centering\arraybackslash}m{1.4cm} >{\centering\arraybackslash}m{1.8cm} >{\centering\arraybackslash}m{2.2cm} }
& mean position error (m) & mean orientation error (degree) & generated successful grasps share (\%) \\
\toprule
bent pipe & 0.004 & 1.94 & 68.2 \\
pulley & 0.005 & 1.32 & 84.2 \\
cinder block & 0.009 & 1.1 & 93.5 \\
\bottomrule
\end{tabular}

%% file: conclusion.tex
This work presents a method to model the grasp space of an underactuated gripper. It generates new gripper configurations that are likely to belong to the grasp space, which allows to explore it. Some insights for a proper hyperparameters tuning are also given.

Various tracks can be investigated in future works. First, a reduction of the number of human inputs required per object would be useful to scale this method to several objects. Moreover, this work was conducted in simulation only, and trials on a real setup should be conducted. Finally, the presented method does not take into account any criterion for the grasp quality. This method and the presented proper hyperparameter tuning can be used to improve other grasp space exploration procedures, which use a grasp quality metric, such as \citep{rolinat_human}.

%% file: root.bbl
\begin{thebibliography}{22}
\providecommand{\natexlab}[1]{#1}
\providecommand{\url}[1]{\texttt{#1}}
\providecommand{\urlprefix}{URL }
\expandafter\ifx\csname urlstyle\endcsname\relax
  \providecommand{\doi}[1]{doi:\discretionary{}{}{}#1}\else
  \providecommand{\doi}{doi:\discretionary{}{}{}\begingroup
  \urlstyle{rm}\Url}\fi

\bibitem[{Abadi et~al.(2015)}]{tensorflow2015-whitepaper}
Abadi, M. et~al. (2015).
\newblock {TensorFlow}: Large-scale machine learning on heterogeneous systems.
\newblock Software available from \url{www.tensorflow.org}.

\bibitem[{{Berenson} et~al.(2007){Berenson}, {Diankov}, {Nishiwaki}, {Kagami},
  and {Kuffner}}]{berenson_grasp_2007}
{Berenson}, D., {Diankov}, R., {Nishiwaki}, K., {Kagami}, S., and {Kuffner}, J.
  (2007).
\newblock Grasp planning in complex scenes.
\newblock In \emph{IEEE-RAS International Conference on Humanoid Robots},
  42--48.

\bibitem[{Choi et~al.(2018)Choi, Schwarting, {DelPreto}, and
  Rus}]{choi_learning_2018}
Choi, C., Schwarting, W., {DelPreto}, J., and Rus, D. (2018).
\newblock Learning object grasping for soft robot hands.
\newblock \emph{{IEEE} Robotics and Automation Letters}, 3(3), 2370--2377.
\newblock \doi{10.1109/LRA.2018.2810544}.

\bibitem[{Chollet et~al.(2015)}]{chollet2015keras}
Chollet, F. et~al. (2015).
\newblock Keras.
\newblock Software available from \url{www.keras.io}.

\bibitem[{Depierre et~al.(2018)Depierre, Dellandréa, and
  Chen}]{depierre_jacquard:_2018}
Depierre, A., Dellandréa, E., and Chen, L. (2018).
\newblock Jacquard: A large scale dataset for robotic grasp detection.
\newblock In \emph{{IEEE}/{RSJ} International Conference on Intelligent Robots
  and Systems ({IROS})}, 3511--3516.
\newblock \doi{10.1109/IROS.2018.8593950}.

\bibitem[{{Drost} et~al.(2010){Drost}, {Ulrich}, {Navab}, and
  {Ilic}}]{drost_model_2010}
{Drost}, B., {Ulrich}, M., {Navab}, N., and {Ilic}, S. (2010).
\newblock Model globally, match locally: Efficient and robust 3d object
  recognition.
\newblock In \emph{IEEE Computer Society Conference on Computer Vision and
  Pattern Recognition}, 998--1005.

\bibitem[{Higgins et~al.(2017)Higgins, Matthey, Pal, Burgess, Glorot,
  Botvinick, Mohamed, and Lerchner}]{Higgins_beta_2017}
Higgins, I., Matthey, L., Pal, A., Burgess, C., Glorot, X., Botvinick, M.,
  Mohamed, S., and Lerchner, A. (2017).
\newblock beta-vae: Learning basic visual concepts with a constrained
  variational framework.
\newblock In \emph{International Conference on Learning Representation}.

\bibitem[{Kingma and Welling(2014)}]{kingma_auto_2014}
Kingma, D. and Welling, M. (2014).
\newblock Auto-encoding variational bayes.
\newblock In \emph{International Conference on Learning Representations
  (ICLR)}.

\bibitem[{Koenig and Howard(2004)}]{koenig_design_2004}
Koenig, N. and Howard, A. (2004).
\newblock Design and use paradigms for gazebo, an open-source multi-robot
  simulator.
\newblock In \emph{{IEEE}/{RSJ} International Conference on Intelligent Robots
  and Systems ({IROS})}, volume~3, 2149--2154.
\newblock \doi{10.1109/IROS.2004.1389727}.

\bibitem[{Levine et~al.(2018)Levine, Pastor, Krizhevsky, Ibarz, and
  Quillen}]{levine_learning_2018}
Levine, S., Pastor, P., Krizhevsky, A., Ibarz, J., and Quillen, D. (2018).
\newblock Learning hand-eye coordination for robotic grasping with deep
  learning and large-scale data collection.
\newblock \emph{The International Journal of Robotics Research}, 37(4),
  421--436.
\newblock \doi{10.1177/0278364917710318}.

\bibitem[{Mahler et~al.(2017)Mahler, Liang, Niyaz, Laskey, Doan, Liu, Ojea, and
  Goldberg}]{mahler_dex-net_2017}
Mahler, J., Liang, J., Niyaz, S., Laskey, M., Doan, R., Liu, X., Ojea, J.A.,
  and Goldberg, K. (2017).
\newblock Dex-net 2.0: Deep learning to plan robust grasps with synthetic point
  clouds and analytic grasp metrics.
\newblock In \emph{Robotics: Science and Systems (RSS)}.

\bibitem[{Pedregosa et~al.(2011)Pedregosa, Varoquaux, Gramfort, Michel,
  Thirion, Grisel, Blondel, Prettenhofer, Weiss, Dubourg, Vanderplas, Passos,
  Cournapeau, Brucher, Perrot, and Duchesnay}]{scikit-learn}
Pedregosa, F., Varoquaux, G., Gramfort, A., Michel, V., Thirion, B., Grisel,
  O., Blondel, M., Prettenhofer, P., Weiss, R., Dubourg, V., Vanderplas, J.,
  Passos, A., Cournapeau, D., Brucher, M., Perrot, M., and Duchesnay, E.
  (2011).
\newblock Scikit-learn: Machine learning in {P}ython.
\newblock \emph{Journal of Machine Learning Research}, 12, 2825--2830.

\bibitem[{Pinto and Gupta(2016)}]{pinto_supersizing_2015}
Pinto, L. and Gupta, A. (2016).
\newblock Supersizing self-supervision: Learning to grasp from 50k tries and
  700 robot hours.
\newblock In \emph{IEEE International Conference on Robotics and Automation
  (ICRA)}, 3406--3413.

\bibitem[{{Roa} et~al.(2008){Roa}, {Suarez}, and {Rosell}}]{roa_grasp_2008}
{Roa}, M.A., {Suarez}, R., and {Rosell}, J. (2008).
\newblock Grasp space generation using sampling and computation of independent
  regions.
\newblock In \emph{IEEE/RSJ International Conference on Intelligent Robots and
  Systems}, 2258--2263.

\bibitem[{Rolinat et~al.(2021)Rolinat, Grossard, Aloui, and
  Godin}]{rolinat_human}
Rolinat, C., Grossard, M., Aloui, S., and Godin, C. (2021).
\newblock Human-initiated grasp space exploration algorithm for an
  underactuated robot gripper using variational autoencoder.
\newblock In \emph{IEEE International Conference on Robotics and Automation
  (ICRA)}, in press.

\bibitem[{Sahbani et~al.(2012)Sahbani, El-Khoury, and
  Bidaud}]{sahbani_overview_2012}
Sahbani, A., El-Khoury, S., and Bidaud, P. (2012).
\newblock An overview of 3d object grasp synthesis algorithms.
\newblock \emph{Robotics and Autonomous Systems}, 60(3), 326--336.

\bibitem[{Santina et~al.(2019)Santina, Arapi, Averta, Damiani, Fiore, Settimi,
  Catalano, Bacciu, Bicchi, and Bianchi}]{santina_learning_2019}
Santina, C.D., Arapi, V., Averta, G., Damiani, F., Fiore, G., Settimi, A.,
  Catalano, M.G., Bacciu, D., Bicchi, A., and Bianchi, M. (2019).
\newblock Learning from humans how to grasp: A data-driven architecture for
  autonomous grasping with anthropomorphic soft hands.
\newblock \emph{{IEEE} Robotics and Automation Letters}, 4(2), 1533--1540.
\newblock \doi{10.1109/LRA.2019.2896485}.

\bibitem[{Scholkopf et~al.(1999)Scholkopf, Smola, and
  Müller}]{Scholkopf_kernel_1999}
Scholkopf, B., Smola, A., and Müller, K.R. (1999).
\newblock Kernel principal component analysis.
\newblock In \emph{Advances in Kernel Methods - Support Vector Learning},
  327--352. MIT Press.

\bibitem[{Sohn et~al.(2015)Sohn, Yan, and Lee}]{sohn_learning_2015}
Sohn, K., Yan, X., and Lee, H. (2015).
\newblock Learning structured output representation using deep conditional
  generative models.
\newblock In \emph{International Conference on Neural Information Processing
  Systems}, volume~2 of \emph{NIPS'15}, 3483--3491. MIT Press, Cambridge, MA,
  USA.

\bibitem[{Townsend(2000)}]{townsend_barretthand_2000}
Townsend, W. (2000).
\newblock The {BarrettHand} grasper – programmably flexible part handling and
  assembly.
\newblock \emph{Industrial Robot}, 27(3), 181--188.
\newblock \doi{10.1108/01439910010371597}.

\bibitem[{{Xue} et~al.(2007){Xue}, {Zoellner}, and {Dillmann}}]{xue_grasp_2007}
{Xue}, Z., {Zoellner}, J.M., and {Dillmann}, R. (2007).
\newblock Grasp planning: Find the contact points.
\newblock In \emph{IEEE International Conference on Robotics and Biomimetics
  (ROBIO)}, 835--840.

\bibitem[{Zhao et~al.(2020)Zhao, Shang, He, and Li}]{zhao_grasp_2020}
Zhao, Z., Shang, W., He, H., and Li, Z. (2020).
\newblock Grasp prediction and evaluation of multi-fingered dexterous hands
  using deep learning.
\newblock \emph{Robotics and Autonomous Systems}, 129, 103550.

\end{thebibliography}
